%% file: main_old.tex
\documentclass[10pt,twocolumn,letterpaper,hyphens]{article}
\pdfoutput=1

\usepackage{cvpr}      % To produce the REVIEW version
\input{additional_package}

\def\cvprPaperID{*****} % *** Enter the CVPR Paper ID here
\def\confName{CVPR}
\def\confYear{2022}

\begin{document}

% \title{Benchmarking the Spatial Robustness in Semantic Segmentation Models: Bridging Natural and Adversarial Localized Corruptions}
\title{Benchmarking the Spatial Robustness of DNNs \\ via Natural and Adversarial Localized Corruptions}

\author{Giulia Marchiori Pietrosanti, Giulio Rossolini, Alessandro Biondi, Giorgio Buttazzo\\
Scuola Superiore Sant'Anna, Pisa\\
Department of Excellence in Robotics \& AI\\
{\tt\small name.surname@santannapisa.it}
}
\maketitle

%%%%%%%%% ABSTRACT
\input{abstract}

%%%%%%%%% BODY TEXT
\input{intro}

\input{rel}

\input{methodology}
\input{experimental}

\input{conclusion}

%%%%%%%%% REFERENCES
{\small
\bibliographystyle{ieee_fullname}
\bibliography{egbib}
}

\end{document}

%% file: additional_package.tex
\usepackage{lipsum}
\usepackage{graphicx}
\usepackage{amsmath}
\usepackage{amssymb}
\usepackage{subcaption}
\usepackage{booktabs}
\usepackage{algorithm}
\usepackage{algpseudocode}
\usepackage{pifont} % For checkbox symbols
\usepackage{array} % For better control of table column width
\usepackage{dsfont}
\usepackage{orcidlink}
\usepackage{enumitem}
\setlist[enumerate]{leftmargin=0mm, label=\alph*)}
\setlist[itemize]{leftmargin=3mm}
\usepackage{hyperref}
\usepackage[table]{xcolor}% http://ctan.org/pkg/xcolor
\usepackage[capitalize]{cleveref}
\crefname{section}{Sec.}{Secs.}
\Crefname{section}{Section}{Sections}
\Crefname{table}{Table}{Tables}
\crefname{table}{Tab.}{Tabs.}
\usepackage{tikz}

\newcommand {\GM}[1]{{\color{black}{#1}}}

%% file: abstract.tex
\begin{abstract}
The robustness of deep neural networks is a crucial factor in safety-critical applications, particularly in complex and dynamic environments (e.g., medical or driving scenarios) where localized corruptions can arise. 
While previous studies have evaluated the robustness of semantic segmentation (SS) models under whole-image natural or adversarial corruptions, a comprehensive investigation into the spatial robustness of dense vision models under localized corruptions remains underexplored.
This paper fills this gap by introducing novel, region-aware metrics for benchmarking the spatial robustness of segmentation models, along with an evaluation framework to assess the impact of natural localized corruptions.
Furthermore, it uncovers the inherent complexity of evaluating worst-case spatial robustness using only a single localized adversarial attack. To address this, the work proposes a region-aware multi-attack adversarial analysis to systematically assess model robustness across specific image regions.
The proposed metrics and analysis were exploited to evaluate 14 segmentation models in driving scenarios, uncovering key insights into the effects of localized corruption in both natural and adversarial forms. The results reveal that models respond to these two types of threats differently; for instance, transformer-based segmentation models demonstrate notable robustness to localized natural corruptions but are highly vulnerable to adversarial ones, and vice versa for CNN-based models. Consequently, we also address the challenge of balancing robustness to both natural and adversarial localized corruptions by means of ensemble models, thereby achieving a broader threat coverage and improved reliability for dense vision tasks.
\end{abstract}

%% file: intro.tex
\section{Introduction}
\label{sec:intro}
In recent years, deep neural networks (DNNs) have demonstrated remarkable performance across various vision applications, increasing their potential applicability in safety-critical domains such as autonomous systems\cite{survey_liu2024survey, PR_QIAN2022108796}. 
Among these scenarios, dense prediction tasks like semantic segmentation (SS) have garnered significant attention. 
These tasks require models to understand the semantic meaning of each pixel in a given scene, often necessitating the extraction of deep contextual information.
In this context, it is crucial to assess the robustness and trustworthiness of SS models, as complex and dynamic environments can give rise to unexpected input variations that challenge the entire system reliability.

The complexity of DNNs and real-world application scenarios, as in autonomous and assisted driving, has made ensuring and evaluating robustness one of the most challenging problems in modern AI \cite{Carlini017,biggio2018wild, Szegedy14, survey_trust}. 
In the literature, robustness evaluation typically involves analyzing model performance under corrupted input conditions, with two main types of corruption commonly studied: \textit{(i) Natural corruptions}, which simulate faults or real-world environmental conditions such as sensor noise or adverse weather (e.g., snow, rain) \cite{hendrycks2018benchmarking}; \textit{(ii) Adversarial corruptions} (adversarial attacks), which involve intentionally crafted perturbations designed to manipulate model predictions. 
While adversarial attacks are often studied as a security threat, they are also a valuable tool for evaluating a model's robustness under a close approximation of worst-case scenarios, contributing to a sense of verifiability \cite{brau2022minimal, survey_trust}.

In recent years, numerous studies have evaluated the robustness of vision models against both natural and adversarial perturbations \cite{hendrycks2018benchmarking, Szegedy14, kamann2020benchmarking}. 
However, a comprehensive characterization of their behavior under \emph{localized} corruptions \cite{patch_object, rossolini_tnnls_2023, saha_role_2020}, which affect only parts of the image, remains insufficiently explored, particularly for dense prediction tasks such as semantic segmentation. 
Indeed, although poorly explored, localized perturbations are highly relevant to real-world scenarios (e.g., sunlight affecting specific regions of an input image) and need to gather more attention in robustness studies.
\GM{Exploring localized corruptions allows addressing critical questions about the \emph{spatial robustness} of semantic segmentation models:}
% This gap raises critical questions about the \emph{spatial robustness} of segmentation models: 
\emph{To what extent do corrupted regions influence predictions in both the corrupted and uncorrupted areas of an image?} \emph{How effectively can models leverage uncorrupted regions to mitigate the impact of corrupted ones?} Addressing these questions is crucial for advancing the robustness and reliability of scene-understanding models in safety-critical environments. 

\noindent\textbf{This paper} addresses the above challenges by studying the effect of localized corruptions across different models through the lens of both \textsc{natural} transformations and \textsc{adversarial} perturbations. 
In particular, the work begins by formalizing the concept of localized corruptions and proposing ad-hoc metrics specifically designed to evaluate spatial robustness in different regions of the input for SS models.

An analysis framework for benchmarking the spatial robustness of SS models against localized natural corruptions is then introduced, enhancing the understanding of the reliability of DNNs in practical scenarios, such as in driving environments, where localized corruptions can occur.

% Next, we investigate critical aspects of adversarial attacks within the context of SS tasks, highlighting that identifying a single localized perturbation capable of comprehensively evaluating spatial robustness (e.g., misclassifying as many pixel predictions as possible) is particularly challenging. This is due to the multi-objective nature of the adversarial optimization problem. To address this challenge, we propose the \emph{region-aware multi-attack adversarial analysis}, a method that iteratively generates localized perturbations, each targeting regions previously unaffected by prior attacks. 
% The purpose of this method is to demonstrate that regions initially considered robust, as they are not misclassified by a single localized attack, can indeed be compromised when specific attack configurations are applied. 
% This insight enhances the understanding of worst-case localized perturbations and their impact on model robustness, highlighting the need to understand the extent to which pixels can be perturbed when a perturbation is applied to a specific area.

Next, critical aspects of adversarial attacks within the context of SS tasks are investigated, highlighting that identifying a single localized perturbation capable of comprehensively evaluating spatial robustness (e.g., misclassifying as many pixel predictions as possible) is particularly challenging. This is due to the multi-objective nature of the adversarial optimization problem. To address this challenge, we propose the \emph{region-aware multi-attack adversarial analysis}, a \GM{novel evaluation algorithm} that iteratively generates localized perturbations, each targeting regions previously unaffected by prior attacks. 
\GM{The idea behind the proposed algorithm builds on the observation that regions initially considered robust (because they are not misclassified by a single localized attack) can, instead, be compromised when repeated runs of the same attack are applied with more targeted configurations. This enables a clearer understanding and evaluation of worst-case localized perturbations, revealing vulnerabilities that may not be sufficiently captured by standard localized attacks.}

The proposed framework, considering both localized natural and adversarial corruptions, is then applied in an experimental evaluation to benchmark the spatial robustness of $14$ SS models in driving scenarios \cite{cordts2016cityscapes}. 
Most interestingly, we observed that \GM{addressed} convolution-based models are significantly vulnerable to localized natural corruptions, while \GM{addressed} transformer-based models exhibit higher robustness. \GM{Conversely, the opposite trend arises when considering localized adversarial corruptions: convolution-based models appear more robust, whereas transformer-based models appear highly vulnerable due to the heavy exploitation of global attention mechanisms.}
These findings raise \GM{contrastive} concerns about the reliability of SS models when evaluated under worst-case safety-related scenarios.
%To address this issue and highlight the importance of achieving a trade-off in terms of spatial robustness between natural transformations and adversarial perturbations, we finally explore the design and configuration of \emph{ensemble} strategies.
\GM{To address this, we finally investigate a promising ensembling analysis aimed at achieving a better trade-off in spatial robustness between natural and adversarial perturbations.}
%, we further provide a preliminary exploration of ensemble strategies that combine models more robust to natural corruptions with those more robust to adversarial perturbations.}

Overall, this study marks a significant step toward understanding localized corruptions and paves the way for developing more robust SS models in practice.\footnote{To support reproducibility, we provide the code and benchmark implementation used in our experiments at the following repository: \url{https://github.com/GiuliaMP/SpatialRobustnessBench}.}

\noindent
In summary, this paper makes the following contributions:
\begin{itemize}
    \item It formalizes new metrics to evaluate the accuracy of SS models while accounting for their spatial robustness under localized corruptions. 
    %Furthermore, we extend existing analysis frameworks to study localized corruption by proposing a more general and flexible analysis framework that allows for quick evaluation of the model robustness.  
    Furthermore, it extends existing approaches to study localized corruption by proposing a flexible framework for evaluating the model spatial robustness.  
    %\item We uncover important insights into the study of localized adversarial perturbations for spatial evaluation in SS models and introduce a new method for generating adversarial perturbations named \textit{region-aware multi-attack adversarial analysis}. 
    \item \GM{It uncovers important insights and challenges in the study of localized adversarial attacks in SS models and introduces a new evaluation method that provides a more thorough assessment of robustness by iteratively testing multiple attacks.}
    %named \textit{region-aware multi-attack adversarial analysis}}. 
    \item The proposed metrics, frameworks, and methods are exploited to analyze the spatial robustness of 14 SS models \GM{on the Cityscape dataset,} under natural and adversarial localized perturbations, revealing contrasting behaviors between the two types of corruptions. Finally, the trade-off between natural and adversarial robustness is \GM{explored} by means of ensemble analysis.
    %, which we explore through the study of ensembling strategies.  
\end{itemize}

The rest of the paper is structured as follows: Section \ref{s:rel} discusses related work and remarks the importance of studying the spatial robustness of dense prediction models; Section \ref{s:methodology} introduces the proposed analysis and formalizes the metrics we used; Section \ref{s:natural} defines the evaluation framework adopted for assessing models under localized natural corruptions; Section \ref{s:adversarial} focuses on localized adversarial perturbations; Section \ref{s:exp} discusses our experimental evaluations and ensemble models, and finally, Section \ref{s:conclusion} states the conclusions and discussion.

%% file: rel.tex
\section{Related Work}
\label{s:rel}
\paragraph{Analysis and benchmarks on DNNs robustness}
Deep learning models are widely used across various applications, including safety-critical domains such as autonomous driving. Given their increasing deployment in such scenarios, assessing their robustness is essential to evaluate their reliability and suitability for real-world use.
To address these challenges, several benchmarks and robustness analyses have been proposed over the years to assess model reliability against both natural and adversarial corruptions. Notable examples include \cite{hendrycks2018benchmarking, croce2022interplay, kamann2020benchmarking, schiappa2024robustness}.

Regarding natural corruptions, the objective is to evaluate models' robustness against perturbations that simulate adverse or potentially out-of-distribution conditions that can be encountered in real-world environments, such as noise, rain, or snow. One of the earliest and most influential studies in this area was conducted by Hendrycks and Dietterich, who introduced a benchmark to assess the resilience of image classification models. This benchmark comprises various natural corruptions, including weather-related disturbances and image formatting artifacts \cite{hendrycks2018benchmarking}.
Since then, these corruptions have been widely adopted and extended in subsequent studies to systematically evaluate and compare the robustness of different models \cite{gu2022vision, bhojanapalli2021understanding, naseer2021intriguing}. 
For instance, Bhojanapalli et al.\ \cite{bhojanapalli2021understanding} investigated the robustness of ViTs comparing it to the robustness of a ResNet-50 model under image corruptions, concluding that ViTs models are at least as robust as the CNN taking into consideration on the majority of the corruptions.
While the majority of studies have focused on classification tasks, fewer works have explored the robustness of models in dense prediction tasks, such as semantic segmentation \cite{schiappa2024robustness, kamann2020benchmarking}, which have highlighted the need for more comprehensive evaluations in this new domain.
% Some recent studies have extended this analysis to benchmark the robustness of transformer-based segmentation models \cite{segformer} \textcolor{blue}{transformer-based SS models like Dino \cite{schiappa2024robustness}}.

Beyond natural corruptions, other works have examined adversarial corruptions \cite{Szegedy14, biggio2018wild}, which represent a worst-case scenario from a security perspective, where perturbations are deliberately crafted to deceive models. Similar to natural corruptions, several studies have compared the robustness of different architectures against adversarial perturbations \cite{bhojanapalli2021understanding, naseer2021intriguing, mahmood2021robustness}. Specifically, research on adversarial robustness in semantic segmentation \cite{yuan2024towards, arnab_robustness_nodate, kamann2020benchmarking} has underscored the importance of establishing more rigorous benchmarks and developing more effective defense strategies to enhance models' robustness in real-world applications.
%C.\ Kamann and C.\ Rother, for example, evaluated different architectural components of DeepLabV3+ under natural corruptions \cite{kamann2020benchmarking}.
%To the best of our knowledge, few works have focused on partially corrupting images with natural corruptions \cite{gu2022vision, guo2023improving}, and no studies have applied a similar approach to semantic segmentation models.
% Some other on adversarial attacks -> gap3) Some of these deepen specifically patch attacks BUT only on ViTs
% In addition to the evaluation of natural corruptions, other works have also faced the use of adversarial corruptions \cite{Szegedy14, biggio2018wild}, which from a safety perspective represents a closer look for a worst case scenario, where intentionally crafted perturbations were designed to deceive models. 
% %Therefore, there has been growing interest in developing defensive strategies and establishing benchmarks to assess model robustness against such perturbations.
% As with natural corruptions, several works have compared the robustness of different models against adversarial perturbations \cite{bhojanapalli2021understanding, naseer2021intriguing, mahmood2021robustness}.
% Some works have specifically analyzed adversarial robustness in semantic segmentation \cite{yuan2024towards, arnab_robustness_nodate, kamann2020benchmarking}, showing the importance of assessing and developing more significant benchmarks and more effective defenses.

\paragraph{Evaluation of localized corruptions and spatial robustness}
While most of the aforementioned studies primarily focus on corruptions affecting the entire input, it is crucial to consider the problem of spatial robustness \cite{saha_role_2020, rossolini_tnnls_2023, gu2022vision}. This aspect is particularly relevant for dense prediction tasks, where model predictions may be influenced by regions beyond the immediate locality of the target area. 
%For instance, in semantic segmentation, pixels that are spatially distant from the region of interest can still impact the model's decision in unexpected ways and vice versa.
\GM{For instance, since semantic segmentation performance is usually evaluated at a pixel level, defining a robustness analysis for such a task opens the possibility to study the impact of perturbed regions on clean regions, and vice versa.}

Despite its importance, research on spatial robustness under natural corruptions remains limited. For example, \cite{gu2022vision} analyzes the robustness of ViTs compared to CNNs under natural and adversarial corruptions, concluding that ViTs generally exhibit greater resilience to natural corruptions than CNNs. However, does not specifically address dense prediction tasks. Conversely, most existing studies on spatial robustness have focused on \GM{physically-realizable adversarial settings, e.g., adversarial patches \cite{pintor_imagenetpatch, brown_adversarial_2018, liu2025_patch, metzen2021meta, rossolini_iccps24}, while only a few works have explored the effects of single localized adversarial perturbations \cite{nakka_indirect_2020}, which instead reveal intrinsic and human-imperceptible weaknesses of the model. Additionally, there is still a lack of comparative analysis on transformer-based models in both cases.}

\GM{Differently from these works, this paper shows that, from an adversarial perspective, a single attack is not sufficient to properly represent the true spatial robustness of dense prediction models. To this end, it proposes an ad hoc evaluation analysis that iteratively adjusts the fooling regions across multiple attacks in the same area, enabling the assessment of regions that conventional attacks might leave unaffected.}
\GM{Furthermore, to address the lack of analysis on natural localized robustness, an aspect particularly relevant for safety assessments, this work introduces a schematic framework and ad hoc metrics to benchmark models in these settings. 

To the best of our knowledge, this work provides the first comprehensive analysis of spatial robustness in the context of both natural and adversarial perturbations for semantic segmentation. This study allowed uncovering important experimental insights when comparing transformer-based and convolution-based architectures and proposing a promising ensembling strategy.}

%% file: methodology.tex
\section{Spatial Robustness Analysis}
\label{s:methodology}
This section first provides some preliminary and background information to theoretically define localized corruptions within the context of SS models. 
Then, new metrics for spatial robustness are proposed to generalize the classic pixel-wise accuracy score used in semantic segmentation.

\subsection{Background and Formalisms}

\paragraph{Preliminaries} 
%%We consider an input image $x \in \mathcal{I}$, where the dimensional space $\mathcal{I}$ has dimensions $H$, $W$, and $C$, representing the height, width, and channels, respectively. 
We consider input images of size $H \times W$ and denote by $\mathcal{I}$ the set of the image's pixels, which are then individually referred to using the index $i$.
A semantic segmentation model designed to classify each pixel into one of $N$ classes is represented by a function $f : x \mapsto \mathbb{R}^{(H \cdot W) \times N}$, which outputs a per-class distribution of scores for each pixel $i$. The vector of output scores for each pixel $i$ is denoted by $f_i$ while its component related to the $j$-th class is denoted by $f^j_i$. 
The semantic label predicted for pixel $i$ is determined by selecting the class with the highest score for that pixel, i.e., $ \operatorname*{argmax}_{j \in \{1, \dots, N\}} f_i^j(x). $
Finally, the corresponding ground-truth segmentation map for the image $x$ is denoted by $y \in \mathbb{N}^{H \times W}$, where each pixel is assigned its true class label, i.e., $y_i \in \{1, \dots, N\}$.

\paragraph{Corruption Areas and Ratio} 
To test the spatial robustness of SS models, we consider corruptions applied to only part of the image. 
Specifically, following the notation adopted in previous work \cite{nakka_indirect_2020, rossolini_tnnls_2023}, we define a \emph{corrupted area} $M$ as a binary mask in the input space $M \in \{0, 1\}^{H \times W}$, which indicates where the corruption is applied. 
For any 2D pixel with index $i$, $M_i = 1$ denotes the pixel is corrupted, while $M_i = 0$ denotes that it is not.
The corrupted image $x_M^c$, given a mask $M$ and a corruption function $c(\cdot)$, is computed as:
\begin{equation}
    x^c_{M} = c(x) \cdot M + x \cdot (1 - M),
\label{eq:corruption_eq}
\end{equation}
\noindent where $c(\cdot)$ can represent either a natural transformation or an adversarial attack. 
Additionally, we introduce the ratio of the perturbed area \GM{as $r = S/(H \cdot W)$, where $\mathcal{S}$ is the number of corrupted pixels, computed from $M$ as $\mathcal{S} = \sum_{i=1}^{H\cdot W} M_i$. It follows that the number $\overline{\mathcal{S}}$ of non-corrupted pixels is $\overline{\mathcal{S}} = \sum_{i=1}^{H\cdot W} 1 - M_i$.}

\subsection{Metrics for Spatial Robustness}
We define metrics for studying the effect of localized corruptions on predictions in different areas of the image, by generalizing both the classic accuracy metrics used for SS and those applied for testing the robustness of image-classification DNNs against corruptions.

\paragraph{Classic Corruption Accuracy} Deriving from the analysis of \cite{hendrycks2018benchmarking} in image classification, given a model \( f \) and a set of possible corruptions \(\mathcal{C}\) (e.g., Gaussian noise, adversarial perturbation or $\{\emptyset\}$ to denote no corruptions), the robustness accuracy of an SS model on a dataset \(\mathcal{D}\) is defined as:
\begin{equation}
A(f, \mathcal{C}) = 
\frac{1}{|\mathcal{C}| \cdot |\mathcal{I}| \cdot |\mathcal{D}|} 
\sum_{c \in \mathcal{C}} 
\sum_{(x, y) \in \mathcal{D}} 
\sum_{i \in \mathcal{I}} \mathds{1} \big(f_i(c(x)) = y_i\big),
\end{equation}

\noindent where \( |\mathcal{C}| \) is the number of corruption types, \( |\mathcal{I}|\) is the number of pixels in the input, \( |\mathcal{D}| \) is the size of the dataset, and $\mathds{1}(\cdot)$ is an operator that returns \lq 1' if a condition is \lq True\rq, otherwise \lq 0\rq. 
Following this metric, we can also define the \emph{Corruption Error} as:
\begin{equation}
    \textit{CE}(f, \mathcal{C}) = \frac{A_{\textit{base}} - A(f, \mathcal{C})}{ A_{\textit{base}}},
    \label{eq:corruptionError}
\end{equation}

\noindent where $A_{\textit{base}}$ refers to the accuracy of a reference model (either $f$ or another one, e.g., AlexNet, as used in \cite{hendrycks2018benchmarking}).
This enables a better understanding of the impact of a given corruption with respect to a reference value.

\paragraph{Localized Corruption Accuracy}
We extend the analysis of the accuracy to consider the case of localized corruption, identified by a corruption mask $M$. 
%For clarity, we denote by \(\mathcal{M}_r\) the set of all masks indicating a corruption ratio $r$. 
%This extends the previous analysis to incorporate the randomness of the corruption mask \( M \), providing deeper insights into how localized corruptions affect model robustness. 
The accuracy of a model $f$ under a corruption $c$ applied in the areas denoted by $M$ is defined as:

\begin{equation}
A_{\mathcal{P}}(f, \mathcal{C}, M)  = 
\frac{1}{|\mathcal{C}| \cdot |\mathcal{D}|} 
\sum_{c \in \mathcal{C}} 
\sum_{(x, y) \in \mathcal{D}} 
\mathcal{K}_{\mathcal{P}}(f, x, c, M, y),
\label{eq:localized_error}
\end{equation}
\noindent with
\begin{equation}
\mathcal{K}_{\mathcal{P}}(f, x, c, M, y) =  \sum_{i \in \mathcal{I}} \mathds{1}\big(f_i(c(x)) = y_i\big) \cdot \mathcal{P}_i(M),
\end{equation}

where \(\mathcal{P}_i(M)\) is a \emph{spatial importance function} designed to weigh the significance of misclassifying a pixel $i$ based on the corruption mask \(M\). 
This function plays a critical role in assessing the impact of regions of the input image, allowing the analysis to either focus on regions of higher relevance or those affected by corruption. 
While, in general, spatial importance functions can be arbitrarily defined, in this study we focus on two specific definitions: 

\begin{itemize}
    \item \textit{Non-Corrupted Region Analysis} (\( A_{\overline{M}}\)): this approach focuses on evaluating the spatial robustness of the model by analyzing regions unaffected by corruptions, uniformly weighting all non-corrupted pixels, i.e.
    \GM{$\mathcal{P}_i(M) = (1 - M_i) /\overline{\mathcal{S}}$.}
    %,where \(\overline{M}\) indicates non-corrupted regions.
    \item \textit{Corrupted Region Analysis (\(A_M\)}): this approach is the dual of the one above, focusing on corrupted pixels only, i.e., \GM{$\mathcal{P}_i(M) = M_i/\mathcal{S}$.}
\end{itemize}

\GM{Notice that both $A_{\overline{M}}$ and $A_M$ are normalized by the number of pixels in their respective regions, ensuring fairness in evaluating and comparing each region. 
%However, the goal is to understand how and where corruptions affect model behavior, rather than to directly compare these values to the clean accuracy. 
Such localized metrics reveal whether the corruption affects predictions beyond its regions, or whether the model can leverage clean informations from outside the perturbed areas to correctly classify the perturbed pixels.}

The first definition (\( A_{\overline{M}}\)) is particularly useful for assessing how well the model remains accurate in non-corrupted regions, offering insights into the capability of corruptions to extend beyond the actual corrupted region. 
 %enables the evaluation of the model's resilience in non-corrupted areas.
Conversely, the second definition ($A_M$) focuses on the model's ability to recover accurate predictions in corrupted regions, potentially leveraging information from non-corrupted regions. 
This analysis is more meaningful when studying natural corruptions, as adversarial attacks can easily lead to complete misclassification of corrupted regions~\cite{rossolini_tnnls_2023, nakka_indirect_2020}, making robustness assessments in such areas less informative.
%
%In our experimental analysis, we highlight that these metrics are expressive for both natural and adversarial corruptions, as they provide measures of resiliency to corruptions within and far from the targeted regions.
% In particular, as better discussed in Section X, for the settings proposed, we evaluate model accuracy against natural corruption by averaging the results over random corrupted regions $M$ selected by $\mathcal{M}_r$, which represents a theoretical set of all masks corresponding to a corruption ratio $r$.
It is worth noting that the metrics presented above can be computed for multiple corruption regions: different settings are considered in our experimental evaluation (Section \ref{s:exp}).
For localized adversarial perturbations, we follow traditional approaches that focus on statically targeting specific areas of the image, such as the center or a corner (e.g., ~\cite{rossolini_tnnls_2023, nakka_indirect_2020}). The rationale behind this choice is that these positions allow evaluating the worst- or best-case scenarios of a model's receptive field \cite{luo2016understanding}. 

Different considerations can be made when assessing the impact of arbitrarily-placed localized corruptions. For natural corruptions, we consider the above metrics while selecting the corruption regions randomly, leading to the following formulation:
\begin{equation}
A_{\mathcal{P}}(f, \mathcal{C})  = 
\frac{1}{|\mathcal{C}| \cdot |\mathcal{D}|} 
\sum_{c \in \mathcal{C}} 
\sum_{(x, y) \in \mathcal{D}} 
\mathcal{K}_{\mathcal{P}}(f, x, c, M_r(x), y),
\label{eq:localized_error_average}
\end{equation}
\noindent where \( M_r(x) \) is a mask that determines the corrupted regions, with patches randomly selected for each input $x$
%selected randomly in each instance 
while respecting the ratio \GM{\( r = S/(H \cdot W)\)} of the corrupted area.

It is finally important to remark that all the proposed metrics focus on pixel-wise accuracy for semantic segmentation. 
Other metrics, such as the mean intersection over union (MIoU), are also frequently used in the field of SS and our analysis can be extended to support them. For instance, no significant modifications are required to support MIoU.

% Please note that the accuracy under localized corruption can be computed across different corruption regions $M$. In particular, for the settings proposed in the experimental part (Section X), we consider the accuracy of the models against natural corruption by averaging the results across random corrupted region $M$ selected by \(\mathcal{M}_r\), which represents a theoretical set of all masks indicating a corruption ratio $r$. 
% Note also that the proposed metrics target the semantic segmentation pixel-wise accuracy, however other standard metrics are often addressed in the field of SS, as the mean-intersection-over-union (MIoU). The analysis prsented above could be promptly extended also to cover MIoU without particular modification of the appraoch. 

\section{Localized \textsc{Natural} Corruptions}
\label{s:natural}
\begin{figure*}[ht]
\includegraphics[width=1.02\textwidth]{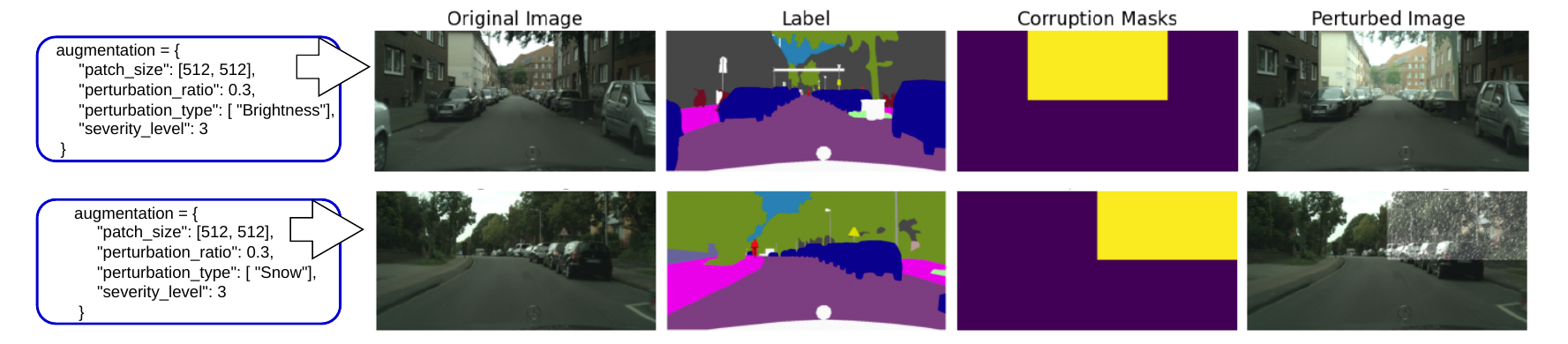}
\centering
%\caption{\small{Illustrations of the proposed framework for evaluating localized natural corruptions. A configuration is specified to define the settings for the localized corruption analysis, and a single sample from the dataset is returned as a tuple consisting of the original image, the original ground truth label, and the corrupted image, along with the corrupted region that highlights which areas have been perturbed.}}
\caption{\small{Illustrations of the proposed framework for evaluating localized natural corruptions. \GM{On the left, two example configurations define the settings for localized corruption analysis: \textit{patch size}, \textit{corruption ratio}, \textit{corruption type}, and the \textit{severity level}. The framework returns a dataset sample as a tuple consisting of the original image, the ground truth label, the corrupted image, and a mask highlighting the corrupted areas.}}}
\label{fig:natural_framework}
\end{figure*}
This section presents the framework designed to evaluate localized natural corruptions.  
\GM{The framework includes all the corruption types available in ImageNet-C \cite{hendrycks2018benchmarking} which, however, does not support localized corruptions.}
% Our framework was inspired by ImageNet-C in \cite{hendrycks2018benchmarking}, which, however, does not support localized corruptions.  
In particular, we consider different transformations \( c \), which can be applied at various severity levels \( s \) (i.e., 1, 2, 3, 4, 5), as selected by the user. 
A natural corruption applied to an input $x$ with severity level $s$ is denoted by \( c(x) = \Gamma(x, s) \), where \( \Gamma \) represents a natural transformation, such as gaussian noise, gaussian blur, motion blur, etc. 
% \GM{The framework includes all perturbations present in ImageNet-C \cite{hendrycks2018benchmarking}. We aim to enable a systematic, reproducible, and broad evaluation of spatial robustness, to reveal model sensitivity to partial corruption across the dense output, and to understand mutual impacts between corrupted and non-corrupted regions.}
%In particular, we allow the analysis of the same corruptions adopted in \cite{hendrycks2018benchmarking}, covering a wide range of robustness and input noise cases.

%To extend the previous analysis, we allow the automatic selection of the
\GM{To enable a systematic, reproducible, and broad evaluation of spatial robustness, the corrupted region indicated by} \( M \) is automatically selected according to the following pipeline.
%, which allows generato generalize the evaluation across different areas. 
The image \( x \) is first divided into non-overlapping patches with size \( \Delta = (\Delta_x, \Delta_y) \), thereby obtaining 
a total of \( P = P_x \times P_y \) patches with 
\( P_x = \lceil W / \Delta_x \rceil \), \( P_y = \lceil H / \Delta_y \rceil \), where \( H \) and \( W \) are the height and width of \( x \), respectively. 
Given a specified corruption ratio \( r \), each patch, and therefore all its pixels, is selected to be perturbed with probability \( r \). 
All pixels within a selected patch are assigned a value of \( 1 \) in \( M \), while the others have a value of \(0\).

In Figure \ref{fig:natural_framework}, we present an illustration of a possible configuration used within the tested framework and the output of a dataset sample in the corrupted validation set of Cityscapes \cite{cordts2016cityscapes}. 
Specifically, a sample corresponds to a tuple containing the original image and its label (kept available to allow comparisons during the evaluation), along with the perturbed image and the corruption region mask \( M \).  
During evaluation, for each image, the set of patches to be perturbed (and thus the mask \( M \)) is recomputed, effectively randomizing the selection of the perturbed areas for the evaluation.

\section{Localized \textsc{Adversarial} Corruptions}
\label{s:adversarial}
This section presents our novel analysis for localized adversarial perturbations. 
Although previous work already addressed the study of localized adversarial perturbations against SS models \cite{nakka_indirect_2020}, we next highlight the difficulties of performing a comprehensive spatial adversarial evaluation using a single attack. 
To this end, we begin by discussing classic approaches and pointing out their drawbacks.
Finally, we introduce our \textit{region-aware multi-attack} algorithm designed to address these issues and offer a better understanding of the worst-case spatial robustness of computer vision models.

\subsection{Attacks for SS from previous work}
In the context of classic adversarial perturbations \cite{Szegedy14, biggio2018wild}, the main idea is to craft a specific adversarial noise \( \delta \) that realizes a worst-case perturbation constrained by a magnitude \( \varepsilon \) under a specific norm, e.g., the \( l_\infty \) norm, such that \( \|\delta\|_\infty \leq \varepsilon \).  
Adversarial perturbations are typically computed by optimizing an adversarial loss function \( \mathcal{L} \), e.g., cross-entropy in image classification, which aims to find a perturbation \( \delta \) for an input $x$ that forces the model to make incorrect predictions:

\begin{equation}
\delta = \arg \max_{\|\delta\|_\infty \leq \varepsilon} \mathcal{L}(f(x + \delta), y),
\label{eq:opt_classic}
\end{equation}

Recalling \Cref{eq:corruption_eq}, it is possible to express inputs affected by localized attacks with a perturbation \( \delta \) and a mask $M$ as follows:
\begin{equation}
x_M^{\delta} = (x + \delta) \cdot M + x \cdot (1 - M).
\label{eq:adv_eq}
\end{equation}

When interested in dense prediction tasks, such as semantic segmentation, where predictions are made at the pixel level, the perturbation is generally computed to maximize the misclassification of all or a subset of pixels in the image \cite{nakka_indirect_2020, rossolini_tnnls_2023}.
In the latter case, it is possible to define a \emph{fooling region} as a subset of the input space for which we intend to induce a misprediction. The fooling region can be arbitrarily defined by a mask $F \in \{0,1\}^{H \times W}$ that does not necessarily need to overlap with the corruption mask $M$.
In this context, the perturbation $\delta$ can be optimized as follows:
\begin{equation}
\delta = \arg \max_{\|\delta\|_\infty \leq \varepsilon} \mathcal{L}_{\mathcal{F}}(f(x^\delta_M), y),
\label{eq:opt_classic_SS}
\end{equation}
\noindent where \( \mathcal{L}_{\mathcal{F}} \) is an adversarial loss designed to focus the model performance in the fooling region $F$ only.
%, and . If \( F \) represents the entire image, i.e., all elements of \( F \) are equal to \( 1 \), then all pixels are targeted by the attack.
% mettere alla fine...qua non serve. 
In practice, to solve the above optimization problem, the perturbation $\delta$ can be iteratively updated with step size \( \alpha \) to maximize the classification error \cite{nakka_indirect_2020, pgd_attack}:
\begin{equation}
\delta_{i+1} \leftarrow \delta_{i} + \alpha \cdot \text{sign} \left( \nabla_{\delta} \mathcal{L}_{\mathcal{F}} \right).
\label{eq:pgd_formulation}
\end{equation}

\subsection{Limitations of previous work}
Although previous approaches are capable of crafting strong perturbations by corrupting as many pixels as possible, we argue that comprehensively evaluating the spatial robustness of semantic segmentation models requires multiple attacks applied to the same area. 
A single perturbation, even when computed using a strong attack, cannot cover all regions susceptible to misprediction. 
This happens because, as shown in Equation~\ref{eq:opt_classic_SS}, crafting an optimal localized adversarial perturbation involves solving a complex multi-objective optimization problem, where the goal is to increase the loss for each pixel in the image, representing a wide range of distinct objectives. 
Many of these objectives are difficult to solve simultaneously, often leading to suboptimal or locally optimal solutions.
To support this observation, we provide representative analysis results in Figures~\ref{fig:inter_class_analysis} and ~\ref{fig:intra_class_analysis} (discussed in the next paragraphs), which highlight, from both inter-class and intra-class perspectives, that adversarial attacks targeting specific subsets of pixels may not generalize effectively to others, making it particularly challenging to craft a single localized perturbation that generates worst-case adversarial effects.

\input{img_methodology_adversarial}

Let us first focus on inter-class attacks with reference to Figure~\ref{fig:inter_class_analysis}, where localized perturbations are applied to the center of the image (\(100 \times 100\), using iterative targeted attacks \cite{nakka_indirect_2020} with \(\epsilon = 32/255\)).
In this example, attacks are performed each targeting different classes in separate runs (\lq road\rq, \lq sky\rq, and \lq car\rq). The outputs of the addressed targeted attacks and their corresponding classes are shown from the third to the fifth subfigures. 
Additionally, we also show the output prediction of an untargeted attack (targeting all classes), in the last subfigure. 
The results demonstrate that the adversarial patterns produce distinct adversarial effects. This highlights that the multi-objective optimization landscape for different target classes is highly divergent, evidencing difficulties for a targeted attack to affect the areas interested by others. 
Furthermore, when observing the output of the untargeted attack, it closely resembles the output of the \lq road' attack, suggesting a bias in the multi-objective optimization process, possibly given by the fact that the majority of pixels of the image are classified as \lq road\rq.
%To further explore this aspect from a different perspective, we also provide an analysis of the embedding similarity between these localized perturbations in the appendix (Sec.~X).

We then analyze the same problem in the context of intra-class attacks. 
In this case, as shown in Figure~\ref{fig:intra_class_analysis}, we crafted attacks to misclassify pixels belonging to the \lq car' class. 
We started with an initial fooling region, highlighted in the first subfigure, consisting of all pixels labeled as \lq car' in the ground truth. After executing the attack of 
Equation~\ref{eq:opt_classic_SS} a first time (denoted by \lq attack 1\rq) using the initial fooling region, we were capable of misclassifying a limited subset of the targeted pixels only (note the attack output in the second row and the corresponding red pixels in the third row of Figure~\ref{fig:intra_class_analysis}).
To carry out a second attack (attack 2), we excluded from the initial fooling area the pixels that were successfully attacked in the previous attack. 
This adjustment allows us to investigate whether previously attacked pixels negatively influence the optimization process for the remaining pixels classified as \lq car\rq. 
The restricted fooling region for the second attack is shown in the third row. 
As can be observed from Figure~\ref{fig:intra_class_analysis}, the second attack expands the attacked area, indicating that previously attacked regions can indeed hinder the optimization process when exploring new attackable regions. 
This process is repeated for a third attack (last columns in the second and third rows of the figure), confirming the previous observation.
%As with the inter-class analysis, an embedding analysis is provided in Sec.~X of the appendix to strengthen these claims.

Finally, for completeness, we present the \textit{cumulative adversarial output} in the third subfigure (first row, third column).
This cumulative output assigns each pixel its adversarial misclassified prediction if at least one attempted attack has misclassified it (the first misclassification is considered for simplicity); otherwise, it retains the original model prediction. 
This approach, provides a clearer worst-case understanding of whether each given pixel can be misclassified by at least one attack and will be used in the algorithm proposed in the next subsection.

\subsection{Achieving Stronger Spatial Adversarial Analysis}
% Following the challenges mentioned in the previous analysis, previously above-mentioned attacks, while aiming to enhance the strength of localized adversarial perturbations by targeting as many pixels as possible, often fail to fully explore the true worst-case spatial robustness of models due to the challenges of addressing multi-objective optimization with a single perturbation. Consequently, single instances of these attacks alone provide an incomplete picture of the model's spatial robustness.
The observations made in the previous section allow concluding that \GM{classically applied} single instances of SS attacks provide an incomplete assessment of the worst-case spatial adversarial robustness of a model.
To address this limitation \GM{and assess a more comprehensive benchmark under localized adversarial perturbations}, a method called \textit{region-aware multi-attack adversarial analysis} is proposed in Algorithm \ref{alg:spatial_robustness}. 
The algorithm builds upon the observations made above and is next explained step-by-step. 
At a high level, the method iteratively generates a new localized adversarial attack on a region $M$ while progressively refining the fooling region $\mathcal{F}$ at each attack run, thereby maximizing the coverage of attacked pixels. 
% At a high level, the method iteratively attacks the input in a specific region while progressively modifying the attacked area by refining the fooling region at each attack run, thereby maximizing the coverage of attacked pixels. 
%The algorithm ensures a systematic evaluation of the model's spatial robustness across the entire image.
% Adversarial perturbation optimization is inherently complex due to the multi-objective nature of fooling multiple regions simultaneously. 
% To address this, the framework begins with a "fooling area" that includes all correctly classified pixels.
% After each attack iteration, misclassified pixels are removed from the optimization landscape, effectively simplifying the problem and narrowing the focus to the remaining correctly classified pixels.

\begin{algorithm}[ht]
\caption{Region-aware Multi-Attack Adversarial Analysis}
\label{alg:spatial_robustness}
\begin{algorithmic}[1]
\Require 
\Statex Model $f$; input $x$; labels $y$; Num. of attacks $N_{\text{Att}}$
%\Ensure Spatial Evaluation of $\hat{y}_{\text{O}}$
\State $\hat{y}_{\text{clean}} \gets \text{argmax}(f(x))$ \Comment{Clean preds}
\State $\hat{y}_{\text{O}} \gets \hat{y}_{\text{clean}}$ \Comment{Initialize cumulative output}
\State $\Psi_{\text{corr}} \gets (\hat{y}_{\text{O}} = y)$ \Comment{Aux.\ mask of correct preds}
\State $\mathcal{F} \gets \Psi_{\text{corr}}$ \Comment{Initial fooling region}
\For{$i \gets 1$ to $N_{\text{Att}}$}
    \State $\mathcal{F} \gets \mathcal{F} \cdot \Psi_{\text{corr}}$ \Comment{Update fooling region}
    \State $x_{\text{adv}} \gets \textit{Attack}_M(x, y, f, \mathcal{F})$\label{line:attack-inst}
    \State $\hat{y}_{\text{adv}} \gets \text{argmax}(f(x_{\text{adv}}))$ \Comment{Preds after attack}
    \State $\Psi_{\text{cur}} \gets (\hat{y}_{\text{adv}} = y)$ \Comment{Current correct pixels}
    \State $\bar{\Psi} \gets (1 - \Psi_{\text{cur}}) \cdot \Psi_{\text{corr}}$ \Comment{Newly misclassified pixels}
    \State $\Psi_{\text{corr}} \gets \Psi_{\text{cur}} \cdot \Psi_{\text{corr}}$ \Comment{Update correct aux. mask}
    \State $\hat{y}_{\text{O}} \gets \Psi_{\text{corr}} \cdot \hat{y}_{\text{O}} + \bar{\Psi} \cdot \hat{y}_{\text{adv}}$ \Comment{Update cumulative output}
\EndFor
\State \Return $\textit{Eval}(\hat{y}_{\text{O}}, y)$
\end{algorithmic}
\end{algorithm}

The proposed algorithm takes as input a given image $x$, its labels $y$, and an SS model $f$, and aims to evaluate the robustness of the image by running multiple attacks ($N_\textit{Att}$), applied to the same area $M$, based on a given method (referred to as $\textit{Attack}_M$ in the algorithm).  
%In particular, the algorithm  is implemented with an untargeted objective to generalize as much as possible, thus covering as many pixels as possible. 
At the beginning of the algorithm, the cumulative output is initialized to the correct model predictions, and the initial fooling region consists of all pixels that are correctly predicted in the non-perturbed output (line 4)\footnote{We address the untargeted case for simplicity and to generalize as much as possible, covering as many pixels as possible. However, the approach can be easily extended to a targeted formulation.}. 
Subsequently, the attack iteratively targets the pixels in the fooling region using a classic iterative attack (e.g., \Cref{eq:pgd_formulation}, line 7), computes the corresponding output (line 8), and updates the fooling region by identifying pixels in the preceding fooling region that have not yet been misclassified in the current attack (lines 9–12 and line 6). 
At the end of each attack, the cumulative adversarial output is updated to replace the predictions of newly misclassified pixels that had previously been correctly classified in earlier attacks.

Finally, the cumulative output is evaluated according to the desired metrics (\emph{Eval} in the algorithm).
It is important to remark that the cumulative output is not derived from the model's prediction on a single perturbed sample but rather from the cumulative worst-case attacked outputs, obtained by aggregating misclassifications across multiple localized perturbations, all concentrated in the same area. 
From a safety perspective, this approach provides a stronger and more accurate understanding of the worst-case robustness of each pixel, demonstrating the existence of at least one perturbation capable of misclassifying them.

% Please note that the proposed formulation is entirely untargeted. As demonstrated in the experimental section, the approach also addresses class-wise coverage by incorporating an incremental fooling region. This strategy helps mitigate intra-class issues that may arise even in targeted attacks, as discussed above and illustrated in Figure~\ref{fig:intra_class_analysis}.
Please note that, while the proposed formulation is entirely untargeted, this strategy also helps mitigate intra-class issues that may arise even in targeted attacks, as discussed above and illustrated in Figure~\ref{fig:intra_class_analysis}.

%% file: img_methodology_adversarial.tex
\newcommand{\captionAdvMethodology}{\small{\GM{Analysis of limitations for single localized perturbations in SS: (a) Inter-class analysis:
clean prediction, attacked image, and predictions across different target classes, shown in sequence. (b) Intra-class analysis: Targeted multi-attack on the class \lq Car\rq. The first row displays the fooling area, the attacked image, and the cumulative adversarial output. The second row shows the output after each attack iteration, while the last row visualizes how the fooling region is selected and which pixels are actually affected at each attack.}}}

\makeatletter
\if@twocolumn
    \begin{figure}[ht]
        \centering
        \begin{subfigure}{\columnwidth}
            \centering
            \includegraphics[width=\textwidth]{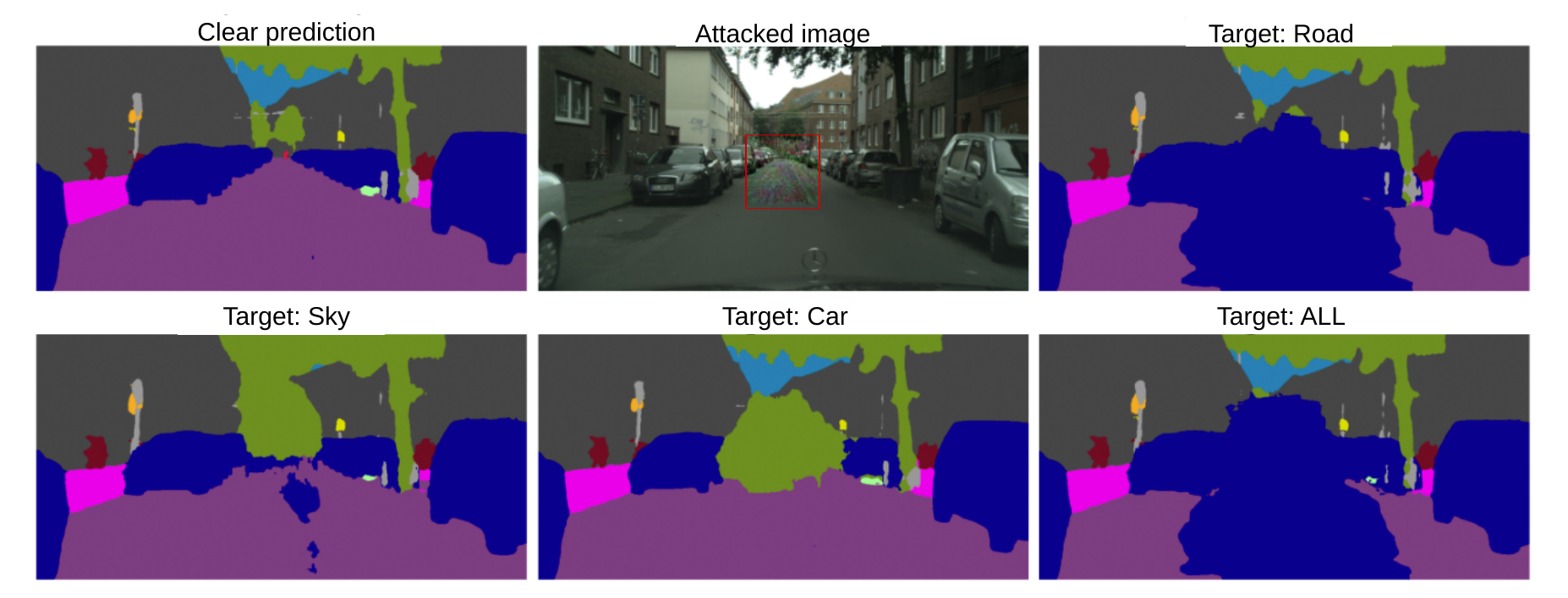}
            \caption{Inter-class target attack analysis.}
            \label{fig:inter_class_analysis}
        \end{subfigure}
        \begin{subfigure}{\columnwidth}
            \centering
            \includegraphics[width=\textwidth]{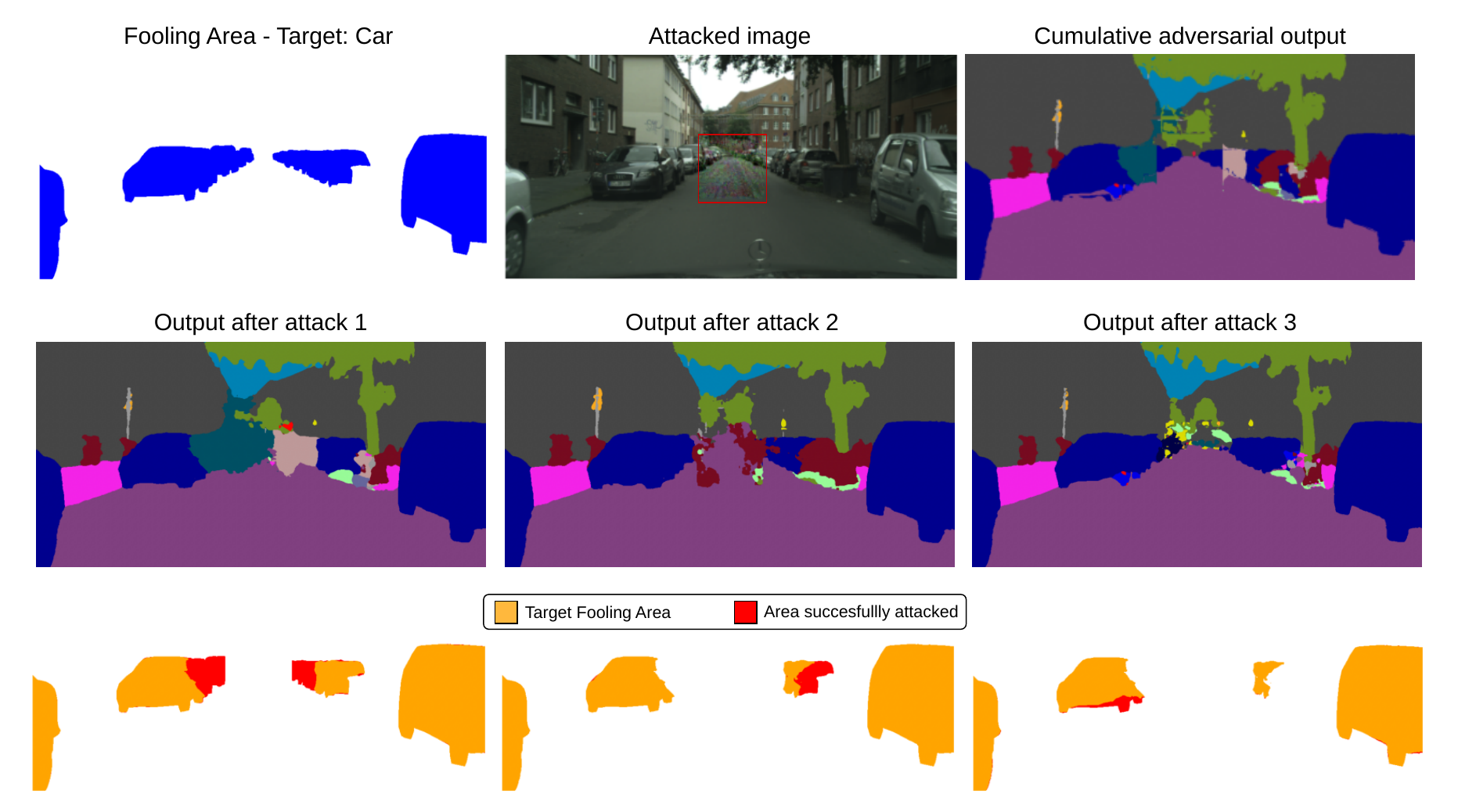}
            \caption{Intra-class target attack analysis}
            \label{fig:intra_class_analysis}
        \end{subfigure}
        \caption{\captionAdvMethodology}
    \end{figure}
\else
    \begin{figure}[ht]
        \centering
        \begin{subfigure}{0.55\columnwidth}
            \centering
            \includegraphics[width=\textwidth]{inter_class_analysis_methodology.pdf.pdf}
            \caption{Inter-class target attack analysis.}
            \label{fig:inter_class_analysis}
        \end{subfigure}
        \begin{subfigure}{0.55\columnwidth}
            \centering
            \includegraphics[width=\textwidth]{intra_class_analysis_methodology.pdf.pdf}
            \caption{Intra-class target attack analysis}
            \label{fig:intra_class_analysis}
        \end{subfigure}
        \caption{\captionAdvMethodology}
    \end{figure}
\fi
\makeatother

%% file: experimental.tex
\section{Evaluation of Spatial Robustness}
\label{s:exp}
In the following, we first describe the experimental setup. 
Then, we evaluate the spatial robustness, based on the proposed metrics, of several models against localized natural corruptions and localized adversarial attacks separately. 
Finally, we discuss the importance of addressing these two aspects together and propose an ensemble-based analysis aimed at balancing localized natural and adversarial robustness.

\subsection{Experimental Setup} 
\label{ss:setup}  
The experiments were conducted with NVIDIA A100 GPUs, using the validation set of Cityscapes \cite{DBLP:conf/cvpr/CordtsORREBFRS16}, which serves as the reference driving dataset for high-resolution scene-understanding segmentation. 
In the context of semantic segmentation, this dataset has frequently been used for robustness evaluation, as it addresses real-world and complex outdoor scenarios \cite{rossolini_tnnls_2023, nakka_indirect_2020, rossolini_iccps24}. 
The Cityscapes validation set includes 500 high-resolution images with an original resolution of ($1024 \times 2048$), resized to manage computational costs (as done in \cite{nakka_indirect_2020}). 
We used this set to evaluate both localized natural corruptions and adversarial attacks.

\noindent\textbf{Models.} We targeted multiple models for semantic segmentation, all known for their ability to achieve high performance on large images while maintaining practical inference times affordable for many real-time applications. 
Specifically, we considered ICNet \cite{icnet_paper}, BiseNet \cite{bisenet_paper}, DDRNet \cite{ddrnet_paper}, PSPNet \cite{pspnet_paper}, SegFormer \cite{xie2021segformer}, PIDNet \cite{xu2023pidnet}, and DeepLabV3 \cite{chen2017rethinking_deeplab}, including multiple versions of each model to evaluate robustness differences, such as variations in the backbone architecture. Pretrained versions, available in the paper's repository, were used for the experimental analysis. The average accuracy of these models on the clean validation set of Cityscapes is reported in the first column of Table \ref{tab:perturbation_analysis}. \GM{For fairness, all results reported for evaluating natural robustness are averaged over 5 runs using the same set of seeds across model comparisons.}
%Standard deviations are omitted for clarity in the plots, as their values were negligible.
%with the other columns discussed in detail in the following sections.

\noindent\textbf{Metrics.} The spatial robustness analysis was conducted by evaluating the pixel-wise accuracy of the models in the \textit{non-corrupted region} $A_{\overline{M}}$ and the \textit{corrupted region} $A_M$, both defined in Section \ref{s:methodology}.
Furthermore, to better understand the robustness degradation as the evaluation parameters vary, we analyzed the previous scores in terms of the \emph{relative corruption error} (RCE) with respect to the accuracy of the addressed models in the same area of interest when no natural corruption or attacks are applied. 
Leveraging the localized corruption accuracy defined in Equation~\ref{eq:localized_error}, the relative corruption error can be computed as

\begin{equation}
    \text{RCE}(f, C, M) = \frac{A_{\mathcal{P}}(f, \{\emptyset\}, M) -  A_{\mathcal{P}}(f, C, M)}{A_{\mathcal{P}}(f, \{\emptyset\}, M)},
    \label{eq:RCE}
\end{equation}

\noindent where \GM{$\mathcal{P}_i(M) = (1 - M_i) /\overline{\mathcal{S}}$.
A similar definition can be derived for the RCE within the corrupted region, considering  {$\mathcal{P}_i(M) = M_i/{\mathcal{S}}$}}.
Note also that, as indicated in Section \ref{s:methodology}, natural corruptions have been evaluated across different regions (see Equation \ref{eq:localized_error_average}).

\subsection{Evaluation of Natural Robustness}  
\input{long_test_natural}  
We first evaluate the natural robustness of the selected SS models. 
Table \ref{tab:perturbation_analysis} reports the results for the \textit{Non-Corrupted Region Analysis} ($A_{\overline{M}}$ in the table) and the \textit{Corrupted Region Analysis} ($A_M$ in the table), considering different localized natural corruptions (synthetic snow, brightness corruptions, and gaussian noise) across different severity levels. For these tests, the corruption ratio is fixed to $r = 0.5$, and corruptions are applied by splitting the images into patches of size $(256, 256)$. 

Considering the robustness in the corrupted region (columns labeled with $A_M$), for snow and brightness adjustments, no model demonstrates a clear and consistent superiority over the others, suggesting that convolution-based architectures remain valid alternatives to SegFormers. 
This observation does not hold in the case of gaussian noise, where transformer-based architectures exhibit a significant accuracy advantage in the corrupted region.

Regarding the robustness in the non-corrupted region (columns labeled with $A_{\overline{M}}$), transformer-like architectures achieve significantly better performance, especially when addressing higher severity levels of localized corruptions.
In these cases, natural corruptions have the potential to substantially affect non-corrupted regions, thereby reducing the accuracy of $A_{\overline{M}}$. 
However, SegFormer demonstrates robustness by preventing the spread of mispredictions to otherwise clean regions.

% As shown in Table \ref{tab:perturbation_analysis}, Segformer architectures achieve often a higher robustness, which is superior robustness compared to convolution-based models in both the $A_{M}$ and $A_{\overline{M}}$ analyses. However, given that several smaller models are particularly relevant for real-time semantic segmentation (SS), it is important to highlight that these models are more vulnerable to pixel mispredictions caused by localized corruptions, as evident in both the IN and OUT analyses.  
% This observation underscores a critical consideration: while lightweight models prioritize efficiency and accuracy, their robustness to localized corruptions—driven by architectural factors—must also be taken into account, especially when applied in safety-critical applications such as autonomous driving.
In Figure \ref{fig:comparisons_natural}, we also show the outputs of the models when localized natural perturbations are applied, while the clean image and corresponding outputs of different models are displayed in the first row. 
For this test, we use a corruption ratio of $r = 0.3$ and a severity level of $3$. The corresponding results align with those in Table \ref{tab:perturbation_analysis} and previous observations, showing that, in general, the SegFormer architectures achieve better robustness both within and outside the corrupted regions.

\begin{figure*}[ht]
    \centering
    % First row of images
    \begin{subfigure}{1.0\textwidth}
        \centering
        \includegraphics[width=0.95\textwidth]{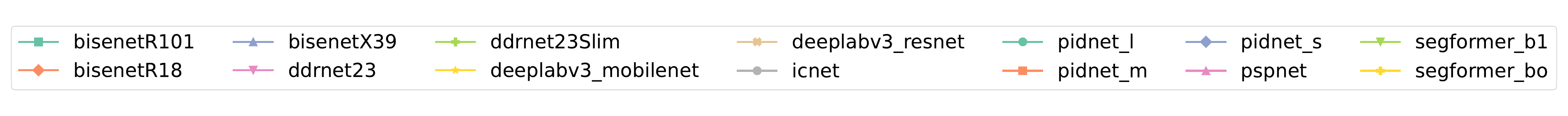}
        \vspace{-0.5em}
    \end{subfigure}
    \begin{subfigure}{1.0\textwidth}
        \centering
        \includegraphics[width=\textwidth]{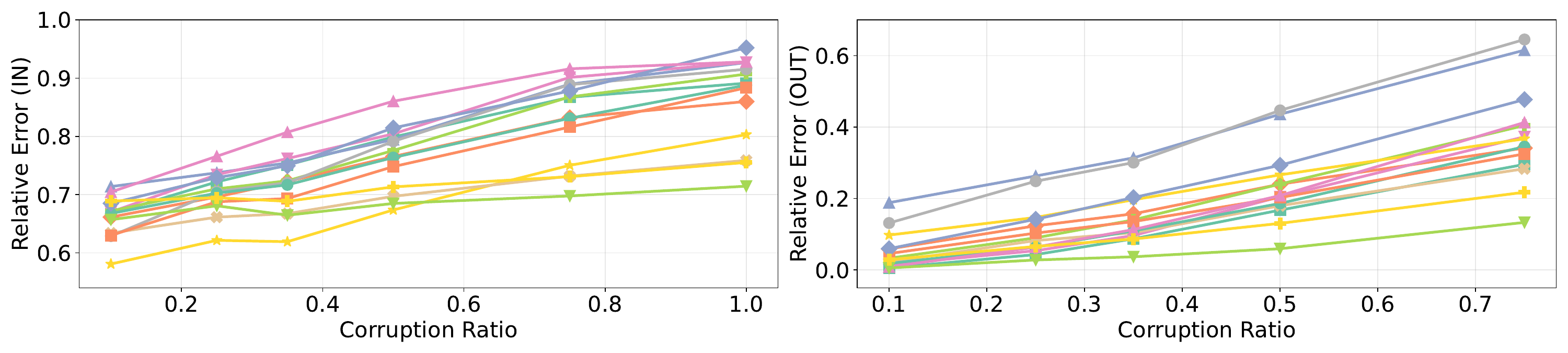}
            \caption{Synthetic snow analysis}
            \label{fig:snow_ratio_comparisons_natural}
    \end{subfigure}
    \begin{subfigure}{1.0\textwidth}
        \centering
        \includegraphics[width=\textwidth]{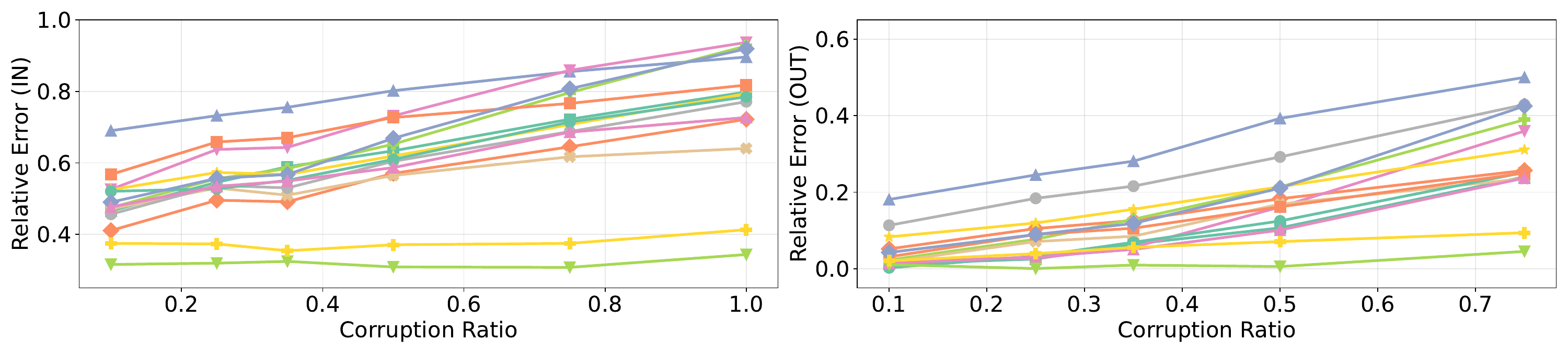}
        \caption{Gaussian noise analysis}
        \label{fig:gaussian_ratio_comparisons_natural}
    \end{subfigure}
    \caption{\small{Analysis of the impact of the corruption ratio. We report the relative \GM{corruption} error \GM{RCE} with respect to the corrupted region (left plots) and the non-corrupted region (right plots). We used synthetic snow and gaussian noise, in the top and bottom plots, respectively, with severity level 3.}}
    \label{fig:ratio_comparisons_natural}
\end{figure*}

\paragraph{Impact of the corruption ratio} To further analyze the impact of localized corruption, we report in Figure \ref{fig:snow_ratio_comparisons_natural} and Figure \ref{fig:gaussian_ratio_comparisons_natural} the variation of the RCE (computed with respect to the clean accuracy in the addressed region) as a function of the 
corruption ratio. 
The RCE was computed both within (left plots) and outside (right plots) the corrupted regions, considering synthetic snow and gaussian noise, respectively.  

As shown in the plots, the corruption error follows different trends across the models. 
For small corruption ratios and corruption types with limited impact, such as synthetic snow, convolution-based models such as DeepLab-MobileNet and DeepLab-ResNet achieve performance comparable to SegFormer, both within and outside the corrupted region. However, when the corruption ratio increases, the error rises more rapidly for the convolution-based models, while SegFormer demonstrates superior robustness. 
As discussed above, a different behavior is observed when considering gaussian noise, for which SegFormer already exhibits significantly higher robustness even for small corrupted regions. 
We believe that this advantage is largely attributed to the application of global attention in transformers, which can effectively attend to uncorrupted areas by leveraging features from those regions or, when regions are completely perturbed, distribute robust features more efficiently across the image \cite{gu2022vision}.

% at the end of the dicussion. 
% To summarize, these results confirm that transformers exhibit greater robustness against natural transformations, even in localized settings, and are better at avoiding mispredictions in non-corrupted areas. This suggests that, even under localized perturbations, transformer-based architectures present a more robust and safer option compared to traditional CNN-based models.

\input{images_adversarial}

\subsection{Evaluation of Adversarial Robustness}
To further understand localized robustness, we evaluate the performance of the SS models using the proposed \textit{region-aware multi-attack adversarial analysis} (Algorithm \ref{alg:spatial_robustness}), aiming for the identification of the worst-case scenario for spatial robustness.  

As discussed in Section \ref{s:adversarial}, for adversarial evaluation, it is important to note that we focus on the analysis of non-corrupted regions ($A_{\overline{M}}$). 
This is because evaluating corrupted regions under localized attacks is straightforward, as any model will typically exhibit near-zero accuracy within the corrupted area.
In contrast, analyzing the non-corrupted regions provides a more meaningful assessment of spatial robustness.  
% \textcolor{blue}{Notice that also in this analysis results are always averaged over 3 seeds.}

\GM{Please note that, for completeness, concerning the adversarial analysis, the full list of parameters involved is: the patch size, the attack magnitude $\epsilon$, the number of attacks \( N_{\text{Att}} \), the position of the attack (which is always centered unless explicitly stated otherwise), and the number of iterations for each attack.}

In Figure \ref{fig:attack_patch_100} and Figure \ref{fig:attack_patch_200}, we report the accuracy in the non-corrupted regions under attack, compared to the non-attacked case (\lq clean\rq, shown in the lightest color in the bars), using localized adversarial attacks with patch sizes of (100, 100) and (200, 200), centered in the image, respectively. 
In both cases, we tested attacks with $\epsilon$ values of $16/255$ and $32/255$ (indicated by different colors in the plots). 
The multi-attack of Algorithm \ref{alg:spatial_robustness} is performed with $N_{\text{Att}}=3$ attacks and 50 optimization iterations (see Equation~\ref{eq:pgd_formulation}) per attack. Further analysis of these parameters is discussed in the following.

%for conclusions.
%These attacks extend the adversarial effect of previous approaches and help characterize the worst-case scenarios produced by adversarial perturbations.  
   
As shown in the plots, the trend is completely opposite to what we observed for natural corruptions. 
In this case, transformers exhibit lower robustness, primarily due to the global attention mechanisms they employ \cite{gu2022vision, xie2021segformer, vaswani_attention_2017}. 
For example, even with a small patch size (100, 100), the accuracy drops significantly from 0.73 to 0.25 and 0.17 for $\epsilon = 16/255$ and $\epsilon = 32/255$, respectively.  
In contrast, convolution-based models, such as DeepLabV3 with ResNet backbone, demonstrate significantly higher robustness. 
This robustness can be attributed to the limited local receptive field inherent to convolutional layers \cite{luo2016understanding}. 
However, it is important to note that these models are not purely convolutional, as they also integrate attention mechanisms to enhance performance. 
This integration can result in substantial accuracy drops, as seen with PSPNet when using patches of size (200, 200). 
This observation is consistent with previous studies, which suggest that even convolutional models employing high-context exploitation through attention-based mechanisms (such as pyramidal squeezing and excitation blocks) can still be vulnerable to certain localized adversarial perturbations \cite{rossolini_tnnls_2023, nakka_indirect_2020}.

\begin{figure}[ht]
    \centering
    % First row of images
    \begin{subfigure}{0.99\columnwidth}
        \centering
        \includegraphics[width=\columnwidth]{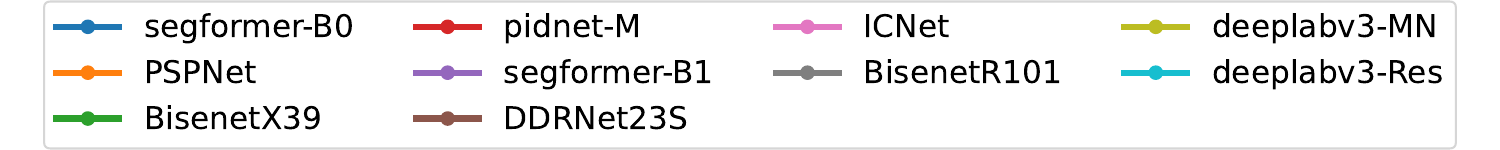}
    \end{subfigure}
    \begin{subfigure}{0.48\columnwidth}
        \centering
        \includegraphics[width=\textwidth]{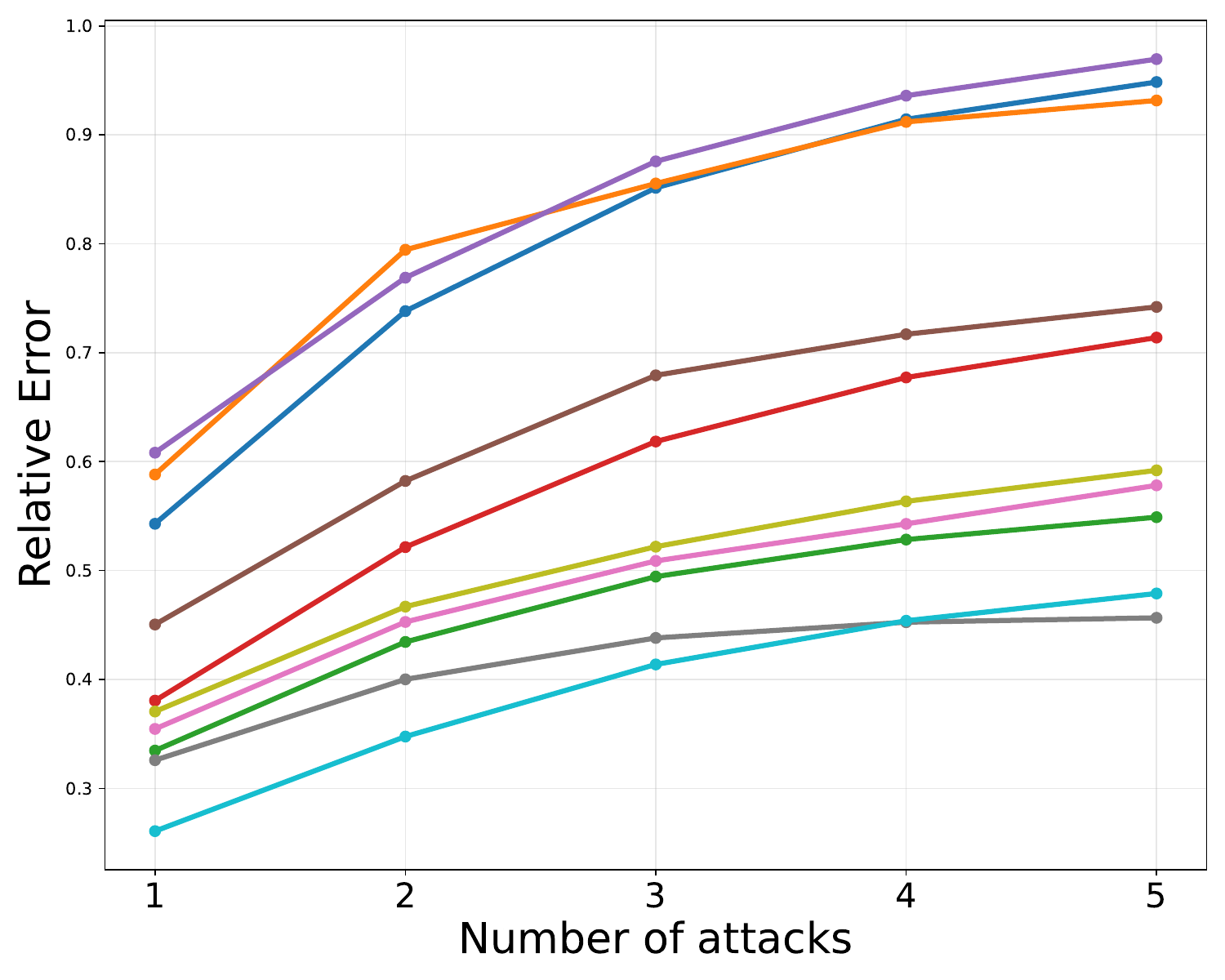}
        \caption{}
        \label{fig:attack_instances}
    \end{subfigure}
    \begin{subfigure}{0.48\columnwidth}
        \centering
        \includegraphics[width=\textwidth]{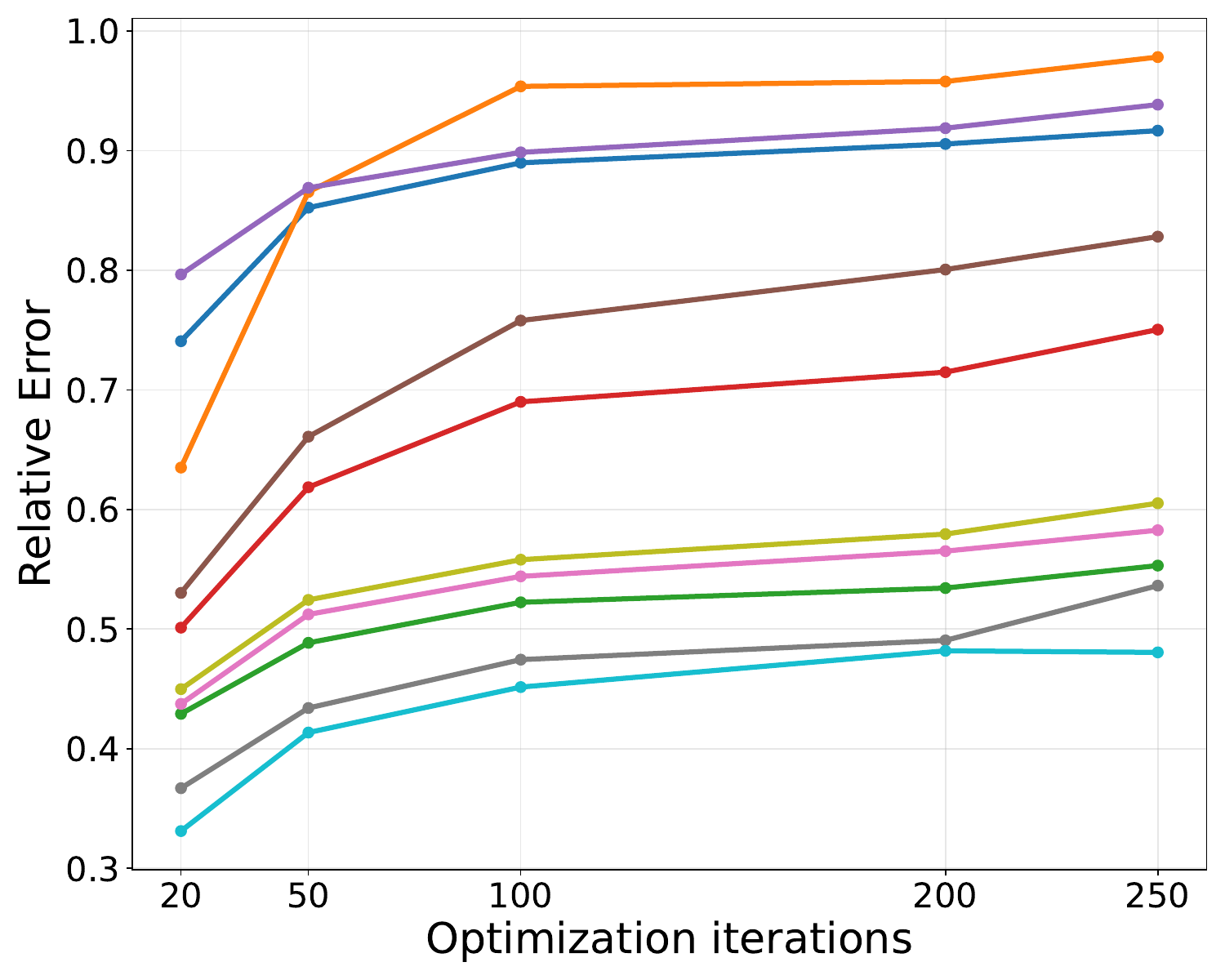}
        \caption{}
        \label{fig:attack_iterations}
    \end{subfigure}
    \caption{\small{\GM{Analysis of the impact of the proposed localized adversarial framework under different attack numbers $N_\textit{Att}$ in Algorithm \ref{alg:spatial_robustness} (a), and under different number of iterations (b). 
    In (a), the number of iterations is fixed to 50, while in (b), the number of attacks is fixed to 3. \GM{In both analyses, patches are of fixed size (200, 200) and $\epsilon = 16/255$}.}}}
    %Analysis of the impact of the proposed localized adversarial analysis when considering multiple attacks $N_\textit{Att}$ in Algorithm \ref{alg:spatial_robustness}(a), with the number of iterations fixed to 50 and $\epsilon = 16/255$ using patches of size (200, 200); and varying the number of optimization iterations per attack in (b), with the number of attacks fixed to 3 and the same settings for $\epsilon$ and patch size.}}
\end{figure}

\paragraph{Benefits of Multi-Attacks}  
To better highlight the benefits provided by Algorithm \ref{alg:spatial_robustness}, we present in Figure \ref{fig:attack_instances} the increase of the RCE, measured with respect to the clean accuracy in the non-corrupted regions, as the number of attacks $N_{\text{Att}}$ in Algorithm \ref{alg:spatial_robustness} increases. 
For these tests, we used patches of size (200, 200), $\epsilon = 16/255$, and 50 optimization iterations per attack.

Specifically, with a single attack (\#1), the multi-attack is equivalent to running a standard attack over the entire targeted region. 
However, as the number of attacks increases, the algorithm begins focusing on fooling regions that have not yet been attacked, thereby expanding the coverage to new regions that were previously unaffected.
In fact, as one may expect, the effectiveness of the multi-attack significantly improves with more attacks, as indicated by the increase in RCE. 
This effect is particularly pronounced in models with high attention mechanisms, such as transformers and PSPNet, demonstrating the importance of the multi-attack approach in identifying worst-case scenarios of localized adversarial perturbations. 
For example, the RCE of SegFormer models increases significantly, rising from approximately 0.6 with a single attack to around 0.95 when using five attacks ($N_{\text{Att}}=5$).

We also provide illustrations of the effect of multiple attacks for SegFormer and PSPNet in Figures \ref{fig:segformer_attack} and \ref{fig:pspnet_attack}. 
As shown in the plots, the adoption of multiple attacks with awareness of the fooling regions allows for overcoming issues in the multi-objective adversarial problem for SS. 
In Figure \ref{fig:segformer_attack}, at each attack, the untargeted multi-attacks target different classes, highlighting the inter-class problem discussed previously in Figure \ref{fig:inter_class_analysis}. 
Similarly, for PSPNet (Figure \ref{fig:pspnet_attack}), addressing different fooling regions leads to a third attack capable of affecting the entire image, thereby overcoming potential objectives that previously limited the effectiveness of earlier attacks.

\paragraph{Impact of the Number of Optimization Iterations}  
We also analyze in Figure \ref{fig:attack_iterations} the impact of varying the number of optimization iterations used at line~\ref{line:attack-inst} in Algorithm \ref{alg:spatial_robustness}, fixing the number of attacks $N_{\text{Att}}=3$. As expected, the RCE increases when using more optimization iterations, such as 200 or 250, highlighting that the complexity of the multi-objective optimization problem in this context requires a higher number of iterations to improve effectiveness.
However, we acknowledge that for large-resolution datasets, as Cityscapes, a limited number of iterations (e.g., 50) is sufficient to provide meaningful and consistent comparisons, as demonstrated in the plots. This balance helps find a trade-off between computational cost and robustness evaluation, particularly when dealing with high-resolution images, which are computationally and memory expensive due to the gradient computations required to update the attack at each step.

\makeatletter
\if@twocolumn
    \begin{figure}[ht]
        \centering
        % First row of images
        \begin{subfigure}{1.\columnwidth}
            \centering
            \includegraphics[width=0.95\columnwidth]{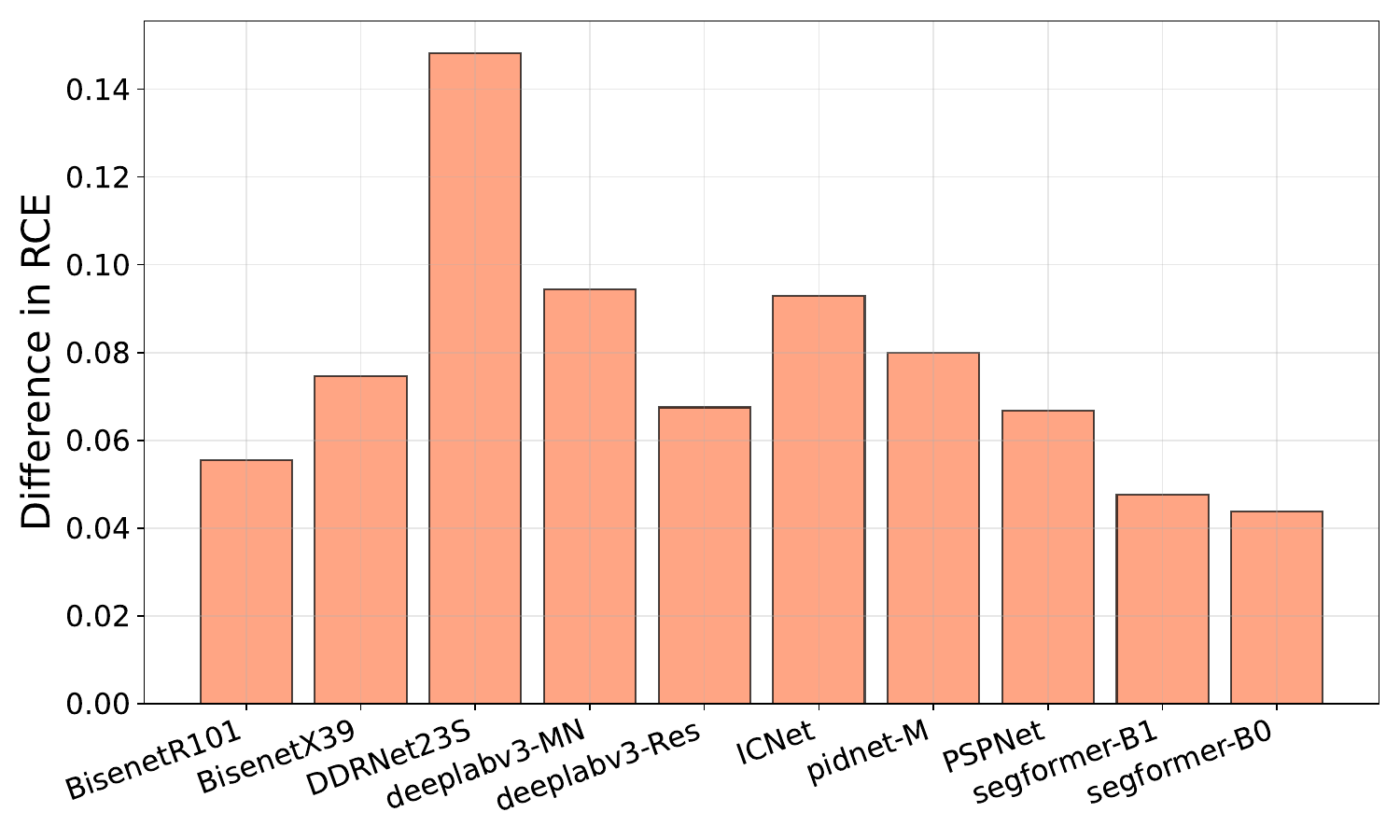}
        \end{subfigure}
        \caption{\small{Differences in relative corruption error when applying localized adversarial attacks at the center of the image versus the bottom-left corner. Patches of size (200, 200) and $\epsilon = 16/255$ are considered.}}
        \label{fig:left_corner_analysis}
    \end{figure}
\else
    \begin{figure}[ht]
        \centering
        % First row of images
        \begin{subfigure}{1.\columnwidth}
            \centering
            \includegraphics[width=0.6\columnwidth]{left_center.pdf}
        \end{subfigure}
        \caption{\small{Differences in relative corruption error when applying localized adversarial attacks at the center of the image versus the bottom-left corner. Patches \GM{are} of \GM{fixed} size (200, 200) and $\epsilon = 16/255$.}}
        \label{fig:left_corner_analysis}
    \end{figure}
\fi
\makeatother

% \begin{figure}[ht!]
%     \centering
%     % First row of images
%     \begin{subfigure}{\columnwidth}
%         \centering
%         \includegraphics[width=\textwidth]{images/methodology/inter_class_analysis_methodology.pdf}
%         \label{fig:inter_class_analysis}
%     \end{subfigure}
%     \caption{Analysis of challenges in multi-objectives adversarial optimization for single localized-pereturbation.}
% \end{figure}

\paragraph{On the Position of the Attacks}  
Finally, we highlight the importance of the position of the adversarial perturbed patch within the image. 
The previous attacks were conducted with the patch placed at the center of the image, which represents the worst-case position for standard convolutional models due to its ability to affect the largest possible receptive field \cite{luo2016understanding, rossolini_iccps24, nakka_indirect_2020}. 
However, it is important to analyze how the effectiveness of localized adversarial attacks decreases when the patch is shifted to more constrained areas, such as the corners of the image.  

In Figure \ref{fig:left_corner_analysis}, we compare the difference between the RCE when attacking the central area versus the one when attacking the bottom-left corner.
%\[
%\text{Diff} = \text{AdvError}_{\text{Center}}%(f) - \text{AdvError}_{\text{Bottom-Left}}(f), 
%\]  
Patches of size (200, 200) and $\epsilon = 16/255$ were considered. 
Specifically, the larger the difference, the more the model demonstrates robustness with respect to the attack position.

As expected, transformer-based architectures, which rely on global attention and process images through patches, exhibit lower sensitivity to the position of adversarial patches. Even when the patch is moved away from the center, the model's response remains close to the worst-case scenario. This highlights the critical risk of attacks for such architectures, being effective from multiple locations within the image, leading to the same adversarial outcome.
%showing similar adversarial effects whether the patch is placed in the center or in other regions. 
%However, the small difference in the effect of patch position is probably also due to the lower robustness of transformer-based architectures to adversarial attacks (i.e., regardless of position, transformer-based architectures consistently exhibit poor performance).
In contrast, convolutional models exhibit more variable behavior depending on the patch’s position.
However, since the tested models are not purely convolutional and incorporate attention mechanisms, their robustness is still partially affected when the patch is placed in a corner. 
For example, BiSeNet with a ResNet-101 backbone and PSPNet demonstrate differences of approximately 0.05 and 0.06 in RCE between central and corner positions, respectively, indicating that localized attacks can still significantly degrade their performance.

\input{ensemble_test}

\subsection{Searching for a Trade-off in Ensembling}  
Given the results obtained in the above analyses of natural perturbations and adversarial attacks, we acknowledge that \GM{the selected} transformer architectures generally exhibit greater robustness to localized natural corruptions by effectively reducing misclassifications, not only within the corrupted regions but also in the surrounding non-corrupted areas. 
However, their reliance on high attention mechanisms makes them more vulnerable to localized adversarial attacks, resulting in a significant drop in accuracy compared to convolutional models tested.  

Following this observation, we emphasize the importance of \textit{searching for a trade-off that balances natural robustness and worst-case adversarial robustness}, which could be particularly beneficial in safety-critical domains such as autonomous driving. 
To address this challenge, we investigate the use of model ensembling, with natural and adversarial localized robustness as key metrics to evaluate the effectiveness of the ensemble strategy.  
To explore this approach, we first define a test-time weighted ensemble strategy \cite{ensemble_GANAIE2022105151} as:  
\begin{equation}
g_{\gamma}(f_1, f_2) = \gamma \cdot f_1 + (1 - \gamma) \cdot f_2,
\label{eq:ensembling}
\end{equation}
where \( f_1 \) and \( f_2 \) are two models selected for the ensemble, and $\gamma$ is a parameter that balances the importance of the softmax scores of model \( f_1 \) with respect to \( f_2 \) in the final model \( g \).  

Specifically, for our tests, we analyze pairs of models where \( f_1 \) is a model with higher robustness to localized adversarial attacks (e.g., DeepLab-ResNet101, DeepLab-MobileNet, BiSeNet-ResNet101, and DDRNet23), and \( f_2 \) demonstrates stronger robustness to natural corruptions but lower robustness to localized adversarial attacks (e.g., SegFormer\_b0 or SegFormer\_b1).

To understand the benefits of ensembling, we redefine the RCE for natural and adversarial perturbations, as follows, to account for the combination of \( f_1 \) and \( f_2 \).
In particular, most important to us for this analysis, we used as metrics: \textit{(i)} the RCE (Equation \ref{eq:RCE}) for natural perturbations within the corrupted area, i.e., 

\begin{equation}
\textit{CE}_\textit{Nat}(g, f_2, C, M) =  \frac{A_\mathcal{P}(f_2, C, M) - A_\mathcal{P}(g, C, M)}{A_\mathcal{P}(f_2, C, M)},
\label{eq:natural_error}
\end{equation}

\noindent where \GM{$\mathcal{P}_i(M) = M_i/\mathcal{S}$} and $C$ is a set of natural corruptions (we considered gaussian noise, with $r = 0.5$); and \textit{(ii)} the RCE for adversarial perturbations outside the corrupted area, i.e.,
\begin{equation}
\quad \textit{CE}_\textit{Adv}(g, f_1, C, M) = \frac{A_\mathcal{P}(f_1, C, M) - A_\mathcal{P}(g, C, M)}{A_\mathcal{P}(f_1, C, M)}, 
\label{eq:adv_error}
\end{equation}
where \GM{$\mathcal{P}_i(M) = (1 - M_i) /\overline{\mathcal{S}}$} and $C$ is a localized adversarial corruption (we used the attack proposed in Algorithm \ref{alg:spatial_robustness}, with $N_\textit{Att} = 5, \epsilon=16/255$). 

The rationale behind these error definitions is that, since \( f_1 \) and \( f_2 \) are chosen as the best-performing models for natural and adversarial robustness, respectively, this formulation allows us to understand how varying \( \gamma \) influences improvements in natural robustness and the corresponding drop in adversarial robustness.  

The results of Figure~\ref{fig:enseble_analysis} confirm the expected trend. As we increase \( \gamma \), the ensemble model becomes more robust to natural perturbations but also increasingly susceptible to localized adversarial attacks. 
Note that, in this case, the attacks are performed directly on the ensemble model to provide a fair evaluation, considering the corresponding \( \gamma \) value.
Interestingly, for certain values (e.g., \( \gamma = 0.4 \)), the adversarial robustness is largely preserved, while we observe a significant reduction in the natural error. 
This phenomenon is particularly noticeable when DeepLab architectures are used as \( f_1 \), as the natural error decreases slightly before the adversarial error curve begins to rise. 
This demonstrates that a positive trade-off between natural and adversarial robustness can be found.

Additionally, we report the mean accuracy of the ensemble on clean inputs in the top portions of the plots to ensure fairness and confirm that there is no drop in clean accuracy. 
The error of clean accuracy is computed with respect to the best-performing model between \( f_1 \) and \( f_2 \). 
As shown in the plots, the use of a well-selected $\gamma$ gets also to negative errors, indicating that the ensemble strategy improves performance on the clean validation set.

We believe this analysis highlights the importance of rethinking model ensembling to achieve results beyond raw performance. 
This approach can be particularly effective when combining architectures with orthogonal strengths, such as robustness to localized natural corruptions and robustness to localized adversarial attacks.

%% file: long_test_natural.tex
\begin{table*}[ht]
\resizebox{1.\textwidth}{!}{
\begin{tabular}{l||r||rr|rr|rr||rr|rr|rr||rr|rr|rr}
\toprule
Type & &\multicolumn{6}{c}{Brightness} & \multicolumn{6}{c}{Snow} & \multicolumn{6}{c}{Gaussian Noise} \\
Severity & Clean &\multicolumn{2}{c}{1} & \multicolumn{2}{c}{3} & \multicolumn{2}{c}{5} & \multicolumn{2}{c}{1} & \multicolumn{2}{c}{3} & \multicolumn{2}{c}{5} & \multicolumn{2}{c}{1} & \multicolumn{2}{c}{3} & \multicolumn{2}{c}{5} \\
Metric & MAcc & $A_M$ & $A_{\overline{M}}$ & $A_M$ & $A_{\overline{M}}$ &  $A_M$ & $A_{\overline{M}}$  &  $A_M$ & $A_{\overline{M}}$ &  $A_M$ & $A_{\overline{M}}$ &  $A_M$ & $A_{\overline{M}}$ &  $A_M$ & $A_{\overline{M}}$ &  $A_M$ & $A_{\overline{M}}$ & $A_M$ & $A_{\overline{M}}$  \\
\midrule
% bisenetX39  & 61.0 & \cellcolor{red!30}0.52 & 0.56 & 0.34 & 0.49 & 0.22 & 0.41  & 0.25 & 0.47 & 0.15 & 0.40 & 0.13 & 0.41 & 0.41 & 0.54 & 0.14 & 0.43 & 0.07 & 0.36 \\
BisenetX39  & 70.6 &  63.1 &  67.0 &47.0 & 61.4 & 34.7 & 54.9 & 34.9 & 57.4 & 22.1 & 51.4 & 20.6 & 51.3 & 50.0 & 61.5 & 20.4 & 50.9 & 12.1 & 43.4 \\
BisenetR18  & 78.0 & 72.6 & 76.0 & 59.5 & 71.5 & 49.1 & 67.9 & 42.2 & 68.7 & 24.7 & 64.2 & 19.7 & 62.0 & 66.4 & 73.9 & 42.4 & 67.6 & 21.5 & 62.4 \\
BisenetR101 & \cellcolor{green!30}82.3 &  \cellcolor{green!30}76.8 & 80.3 & 63.7 & 77.3 & 53.7 & \cellcolor{green!30} 74.1 & 39.7 & 72.5 & 20.8 & 68.2 & 19.8 & 68.2 & 64.0 & 78.5 & 35.8 & 74.3 & 22.4 & 70.7 \\
DDRnet23 & 83.2 & \cellcolor{green!30}78.7 & \cellcolor{green!30}81.7 & 64.7 & 76.4 & 51.2 & 72.5 & 39.2 & 71.8 & 22.5 & 67.4 & 18.5 & 65.1 & 64.2 & 78.1 & 27.3 & 67.9 & 13.7 & 59.8 \\
DDRnet23Slim & 80.8  & 75.0 & 78.9 & 61.4 & 73.0 & 49.4 & 68.8 & 39.9 & 68.2 & 25.4 & 63.5 & 20.1 & 60.4 & 63.5 & 74.1 & 32.9 & 64.4 & 15.0 & 58.3 \\
Icnet & 72.7 & 65.8 & 70.4 & 50.3 & 64.0 & 41.4 & 58.6 & 40.9 & 58.7 & 22.7 & 48.1 & 20.4 & 47.6 & 59.2 & 67.3 & 37.5 & 58.1 & 20.8 & 47.8 \\
PSPnet & 81.6 & 75.1 & 79.1 & 60.5 & 74.5 & 49.1 & 71.6 & 35.2 & 70.1 & 14.6 & 63.6 & 11.6 & 60.4 & 66.8 & 78.2 & 35.3 & 70.9 & 15.6 & 59.8 \\
Segformer-bo &79.3 & 75.8 & 78.2 & 67.6 & 76.4 & 62.0 & 75.7 & 45.9 & \cellcolor{green!30}73.2 & 26.5 & \cellcolor{green!30}70.9 & 19.6 & \cellcolor{green!30}69.9 & \cellcolor{green!30}68.8 & 77.1 & \cellcolor{green!30}54.0 & \cellcolor{green!30}74.8 & \cellcolor{green!30}34.8 & \cellcolor{green!30}73.0 \\
Segformer-b1 & 80.9 & 77.4 & 80.2 & \cellcolor{green!30}70.2 & \cellcolor{green!30}79.1 & \cellcolor{green!30}66.0 & \cellcolor{green!30}78.6 & \cellcolor{green!30}50.1 &  \cellcolor{green!30}77.3 & 27.9 &  \cellcolor{green!30}75.4 & 21.1 &  \cellcolor{green!30}74.5 & \cellcolor{green!30}72.5 & \cellcolor{green!30}80.2 & \cellcolor{green!30}60.2 & \cellcolor{green!30}79.1 & \cellcolor{green!30}42.2 & \cellcolor{green!30}77.7 \\
Deeplabv3-m & 76.7 & 73.3 & 75.0 & 66.6 & 72.3 & 59.2 & 69.4 & \cellcolor{green!30}50.5 & 69.8 &  \cellcolor{green!30}32.1 & 63.9 &  \cellcolor{green!30}25.2 & 59.9 & 56.4 & 71.4 & 36.3 & 64.4 & 20.2 & 57.6 \\
Deeplabv3-r & 80.6 &  77.0 & 78.6 & \cellcolor{green!30}71.7 & 75.7 & \cellcolor{green!30}64.8 & 73.0 & 49.3 & 72.6 & \cellcolor{green!30}29.2 & 68.4 & \cellcolor{green!30}22.3 & 66.1 & 65.8 & 76.4 & 40.3 & 67.6 & 16.3 & 58.1 \\
Pidnet-s & 80.0 & 74.1 & 77.1 & 61.7 & 73.4 & 48.8 & 69.0 & 41.7 & 69.5 & 22.3 & 63.9 & 18.2 & 61.2 & 65.6 & 76.3 & 34.3 & 66.2 & 12.2 & 52.8 \\
Pidnet-m & 80.6 & 75.9 & 79.2 & 62.9 & 75.2 & 48.9 & 71.2 & 47.6 & 73.5 & 28.1 & 69.6 & 22.6 & 67.1 & 59.4 & 75.2 & 27.3 & 68.8 & 10.1 & 64.4 \\
Pidnet-l & \cellcolor{green!30}82.1 & 77.2 & \cellcolor{green!30}81.1 & 60.9 & \cellcolor{green!30}76.7 & 47.3 & 72.4 & 47.7 & 74.3 & 24.9 & 69.3 & 19.0 & 67.4 & 69.0 & \cellcolor{green!30}79.5 & 36.4 & 71.7 & 15.6 & 63.3 \\
\bottomrule
\end{tabular}
}
\caption{\small{\GM{Results of the analysis for} natural corruptions considering brightness, snow, and gaussian noise as corruption types at different severity levels (1, 2, 3). The first column reports the clean average accuracy (\emph{MAcc}), while the remaining columns analyze the accuracy within and outside the corrupted region, represented as $A_M$ and $A_{\overline{M}}$, respectively, following the analysis presented in Section \ref{s:methodology}. The corruption ratio is fixed at 50\% of the image. For each metric and analysis, the top-2 models are highlighted in green for easier interpretation.}}
\label{tab:perturbation_analysis}
\end{table*}

\begin{figure*}[ht]
    \centering
    % First row of images
    \begin{subfigure}{\textwidth}
        \centering
        \includegraphics[width=\textwidth, trim=0 0 0 0.0cm,, clip]{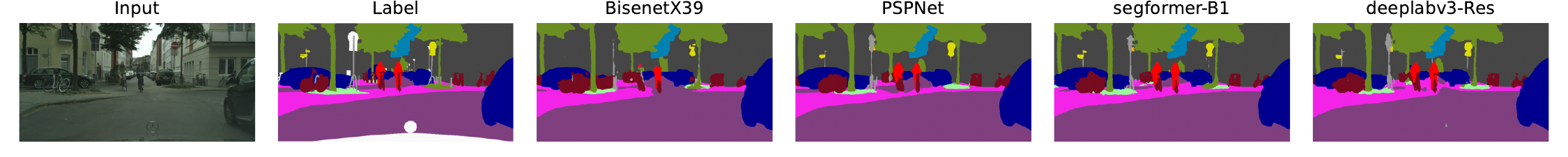}
    \end{subfigure}
    \begin{subfigure}{\textwidth}
        \centering
        \includegraphics[width=\textwidth, trim=0 0 0 0.0cm,, clip]{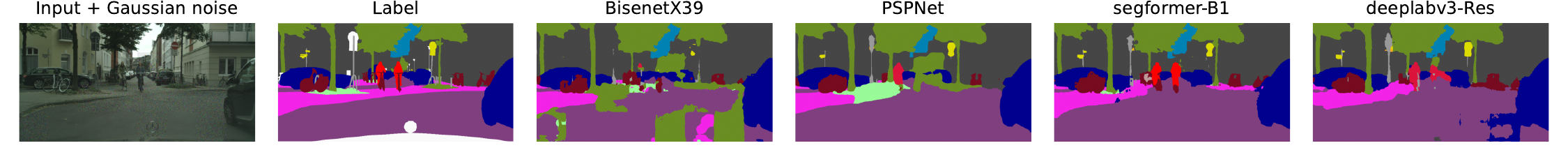}
    \end{subfigure}
    \begin{subfigure}{\textwidth}
        \centering
        \includegraphics[width=\textwidth,  trim=0 0 0 0.0cm, clip]{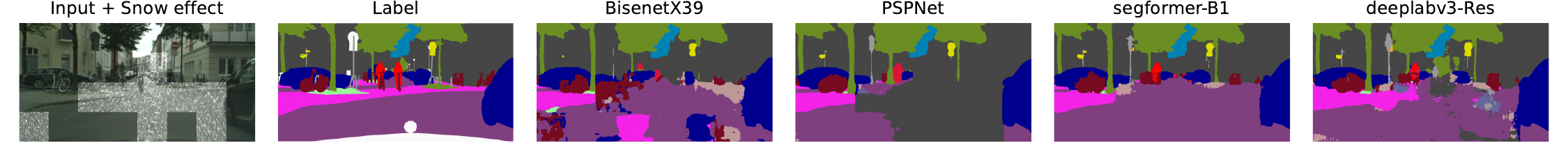}
    \end{subfigure}
    \caption{\small{Illustrations of the effects of localized natural corruptions across different semantic segmentation models (BiseNetX39, PSPNet, SegFormer-B1, DeepLabV3-ResNet) with a perturbation ratio of $r=0.3$ and severity level 3. \GM{The first row shows the clean image along with its ground truth label and corresponding model outputs. The second row presents the same image after applying localized Gaussian noise, and the third row shows it with synthetic snow applied.}}}
    \label{fig:comparisons_natural}
\end{figure*}

%% file: images_adversarial.tex
\newcommand{\captionAdvPlots}{\small{
\GM{Analysis of the accuracy in the non-corrupted region with (dark color) and without (light colors) localized adversarial perturbations. Attacks were performed using $\epsilon = 16/255$ and $\epsilon = 32/255$.}
%Accuracy in the non-corrupted region with \GM{(dark color)} and without (light color) localized adversarial perturbations using $\epsilon = 16/255$ and $\epsilon = 32/255$. 
The patch is always applied at the center of the image, with a size of (100, 100) in (a) and (200, 200) in (b), respectively.}}
\newcommand{\captionAdvIllustrations}{\small{Illustration of the impact of using the \textit{Region-aware multi-attack adversarial analysis} (Algorithm \ref{alg:spatial_robustness}) on Segformer (a) and PSPNet (b) for two different images in the Cityscapes validation set. 
%\GM{Clean image, first attack iteration, and ground truth label are shown in the first row of both (a) and (b), while the second row shows the cumulative output of each model.}
The use of multiple attacks allows for coverage of different parts of the input pixels, highlighting the complexity of solving a multi-objective adversarial optimization problem for SS and demonstrating how the proposed algorithm can overcome this challenge. The attacks shown here use patches of size (200, 200) with $\epsilon = 32/255$.}}

\makeatletter
\if@twocolumn
    \begin{figure}[ht]
    \centering
    % First row of images
    \begin{subfigure}{0.98\columnwidth}
        \centering
        \includegraphics[width=\columnwidth]{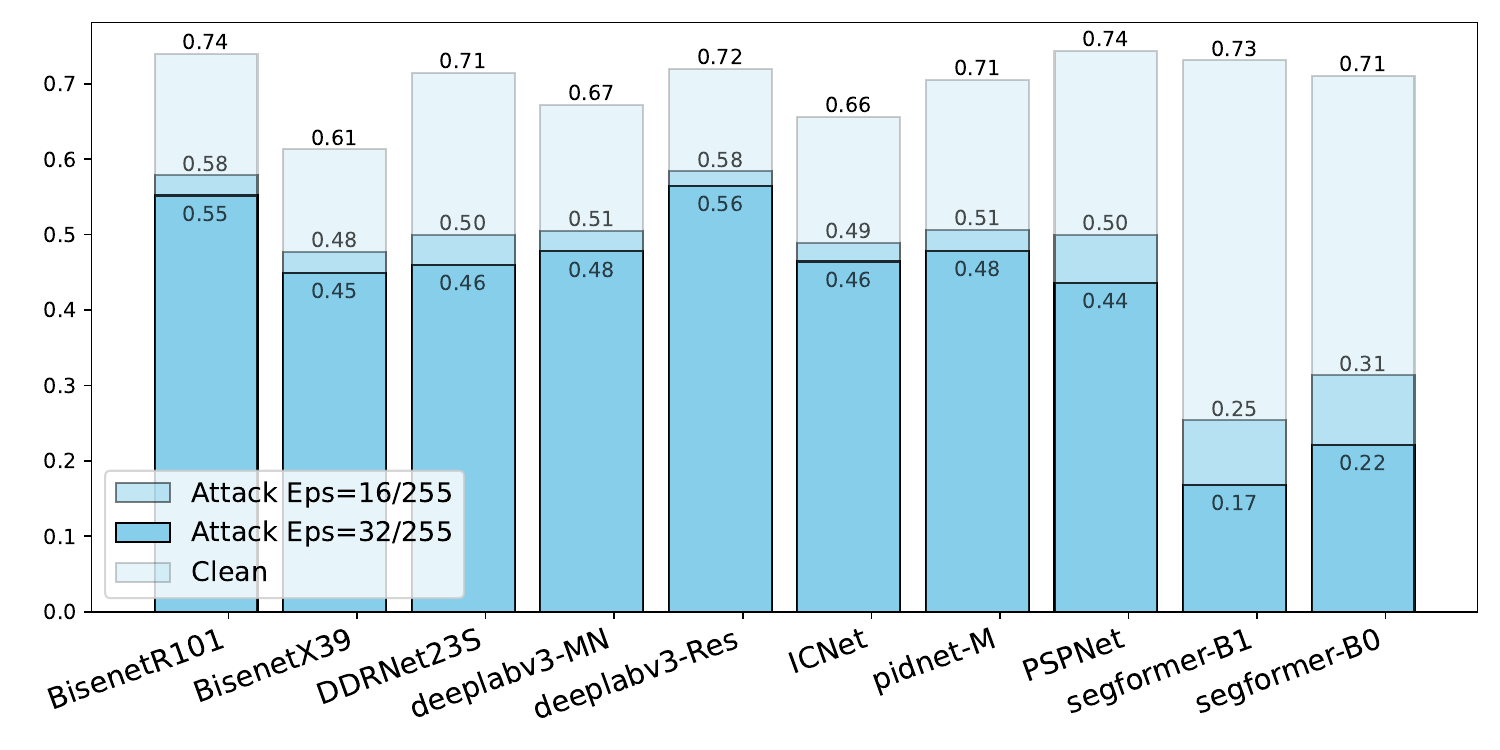}
        \caption{Attack with patches (100,100)}
        \label{fig:attack_patch_100}
    \end{subfigure}
    \begin{subfigure}{0.98\columnwidth}
        \centering
        \includegraphics[width=\columnwidth]{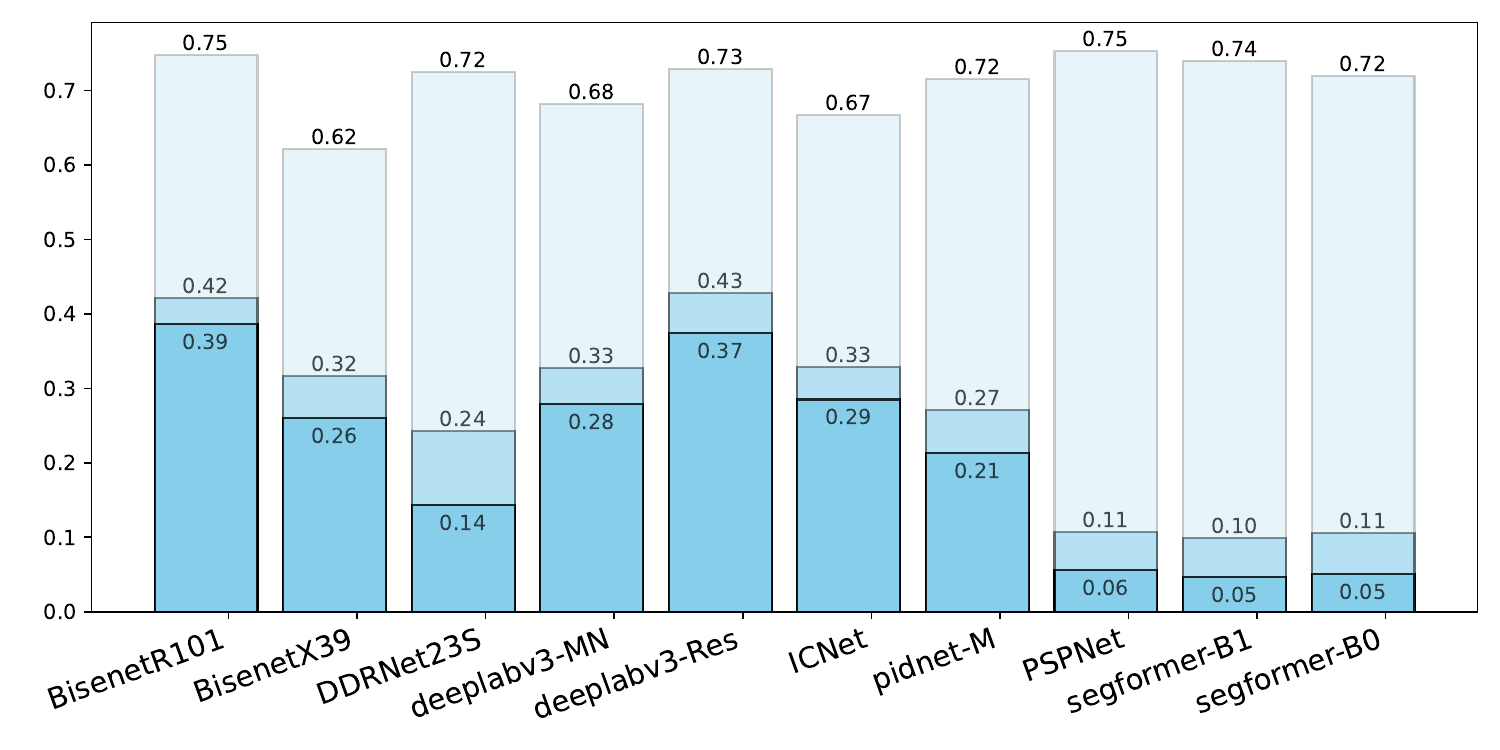}
        \caption{Attack with patches (200,200)}
        \label{fig:attack_patch_200}
    \end{subfigure}
    \caption{\captionAdvPlots}
    \end{figure}
    \begin{figure}[ht]
    \centering
    % First row of images
    \begin{subfigure}{0.98\columnwidth}
        \begin{subfigure}{\textwidth}
        \centering
        \includegraphics[width=\textwidth]{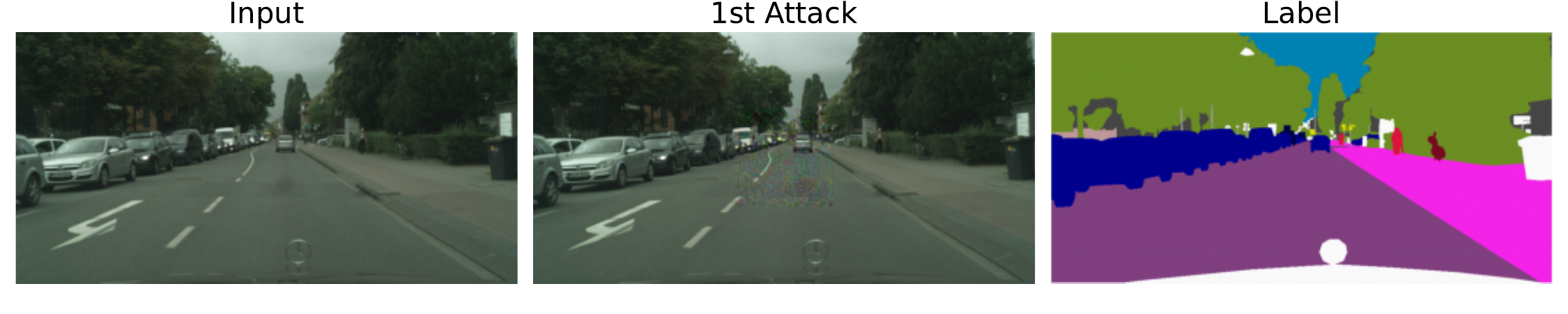}
        %\label{fig:inter_class_analysis}
        \end{subfigure}
        \begin{subfigure}{\textwidth}
        \centering
        \includegraphics[width=\textwidth]{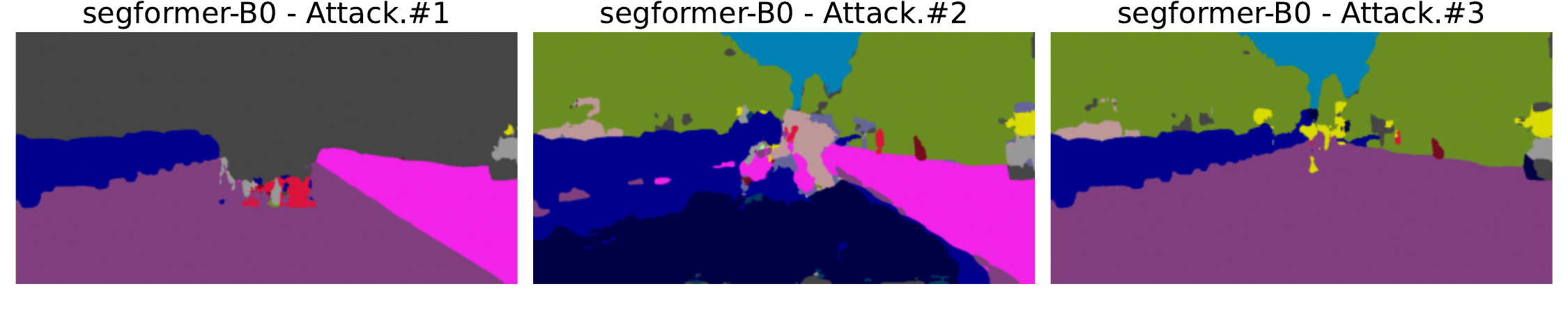}
        %\label{fig:inter_class_analysis}
        \caption{Segformer Attack}
        \label{fig:segformer_attack}
        \end{subfigure}
    \end{subfigure}
    \begin{subfigure}{0.98\columnwidth}
        \begin{subfigure}{\textwidth}
        \centering
        \includegraphics[width=\textwidth]{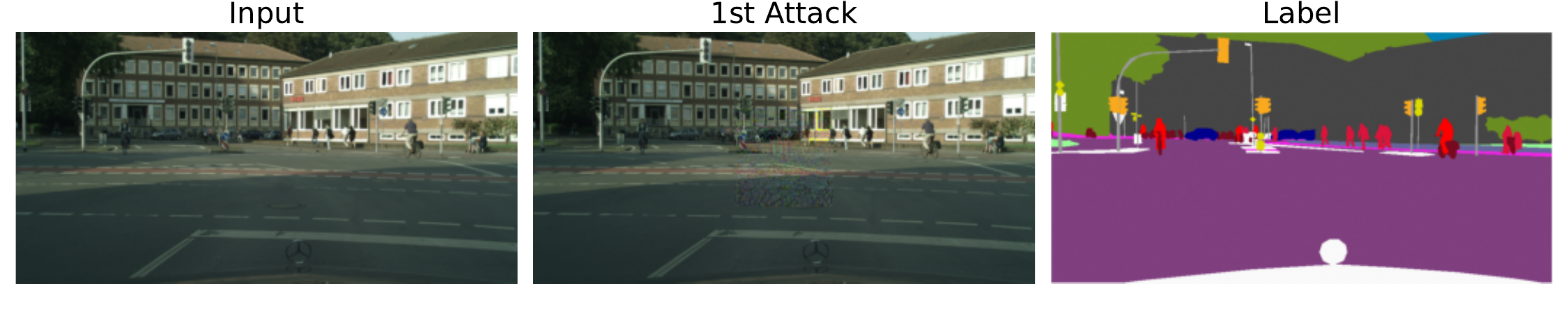}
        %\label{fig:inter_class_analysis}
        \end{subfigure}
        \begin{subfigure}{\textwidth}
        \centering
        \includegraphics[width=\textwidth]{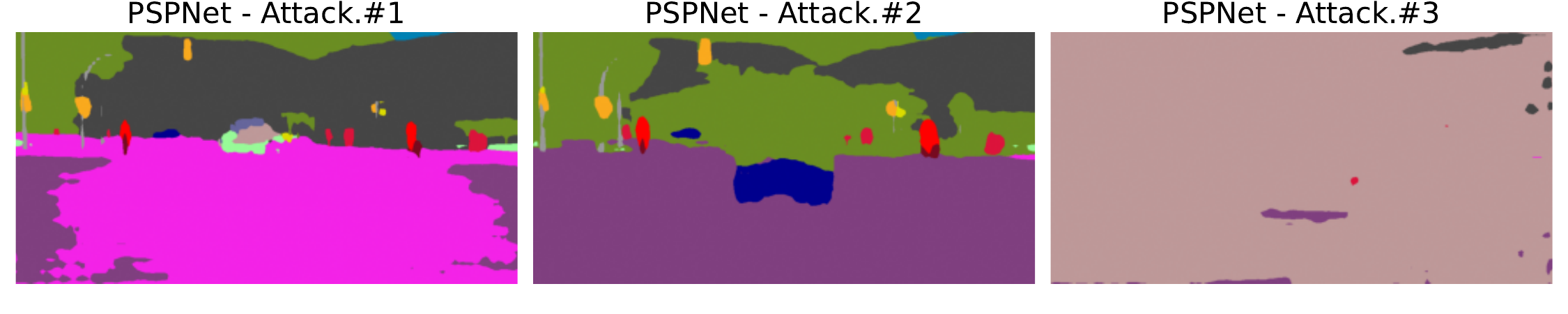}
        \caption{PSPNet Attack}
        \label{fig:pspnet_attack}
        \end{subfigure}
    \end{subfigure}
    \caption{\captionAdvIllustrations}
    \end{figure}
\else
    \begin{figure*}[ht]
    \centering
    % First row of images
    \begin{subfigure}{0.72\textwidth}
        \centering
        \includegraphics[width=\textwidth]{100_attack_general.pdf}
        \caption{Attack with patches (100,100)}
        \label{fig:attack_patch_100}
    \end{subfigure}
    \begin{subfigure}{0.72\textwidth}
        \centering
        \includegraphics[width=\textwidth]{200_attack_general.pdf}
        \caption{Attack with patches (200,200)}
        \label{fig:attack_patch_200}
    \end{subfigure}
    \caption{\captionAdvPlots}
\end{figure*}

\begin{figure*}[ht]
    \centering
    % First row of images
    \begin{subfigure}{0.75\columnwidth}
        \begin{subfigure}{\textwidth}
        \centering
        \includegraphics[width=\textwidth]{segformer_images_attack.pdf}
        %\label{fig:inter_class_analysis}
        \end{subfigure}
        \begin{subfigure}{\textwidth}
        \centering
        \includegraphics[width=\textwidth]{segformer_attacks.pdf}
        %\label{fig:inter_class_analysis}
        \caption{Segformer Attack}
        \label{fig:segformer_attack}
        \end{subfigure}
    \end{subfigure}
    \begin{subfigure}{0.75\columnwidth}
        \begin{subfigure}{\textwidth}
        \centering
        \includegraphics[width=\textwidth]{pspnet_images_attack.pdf}
        %\label{fig:inter_class_analysis}
        \end{subfigure}
        \begin{subfigure}{\textwidth}
        \centering
        \includegraphics[width=\textwidth]{pspnet_attacks.pdf}
        \caption{PSPNet Attack}
        \label{fig:pspnet_attack}
        \end{subfigure}
    \end{subfigure}
    \caption{\captionAdvIllustrations}
\end{figure*}
\fi
\makeatother

%% file: ensemble_test.tex
\newcommand{\captionEnsembleTest}{\small{Analysis of the natural and adversarial errors, computed as discussed in Equations \ref{eq:natural_error} and \ref{eq:adv_error}, respectively, and clean accuracy when considering different models as \( f_1 \) and \( f_2 \) in the ensemble strategy defined in Equation \ref{eq:ensembling}, for different configurations of \( \gamma \). The results aim to explore the existence of a possible trade-off to balance robustness against localized natural corruptions and adversarial perturbations.}}

\makeatletter
\if@twocolumn
    \begin{figure*}[ht]
        \centering
        % First row of images
        \begin{subfigure}{\textwidth}
        \begin{subfigure}{0.24\textwidth}
            \centering
            \includegraphics[width=\textwidth]{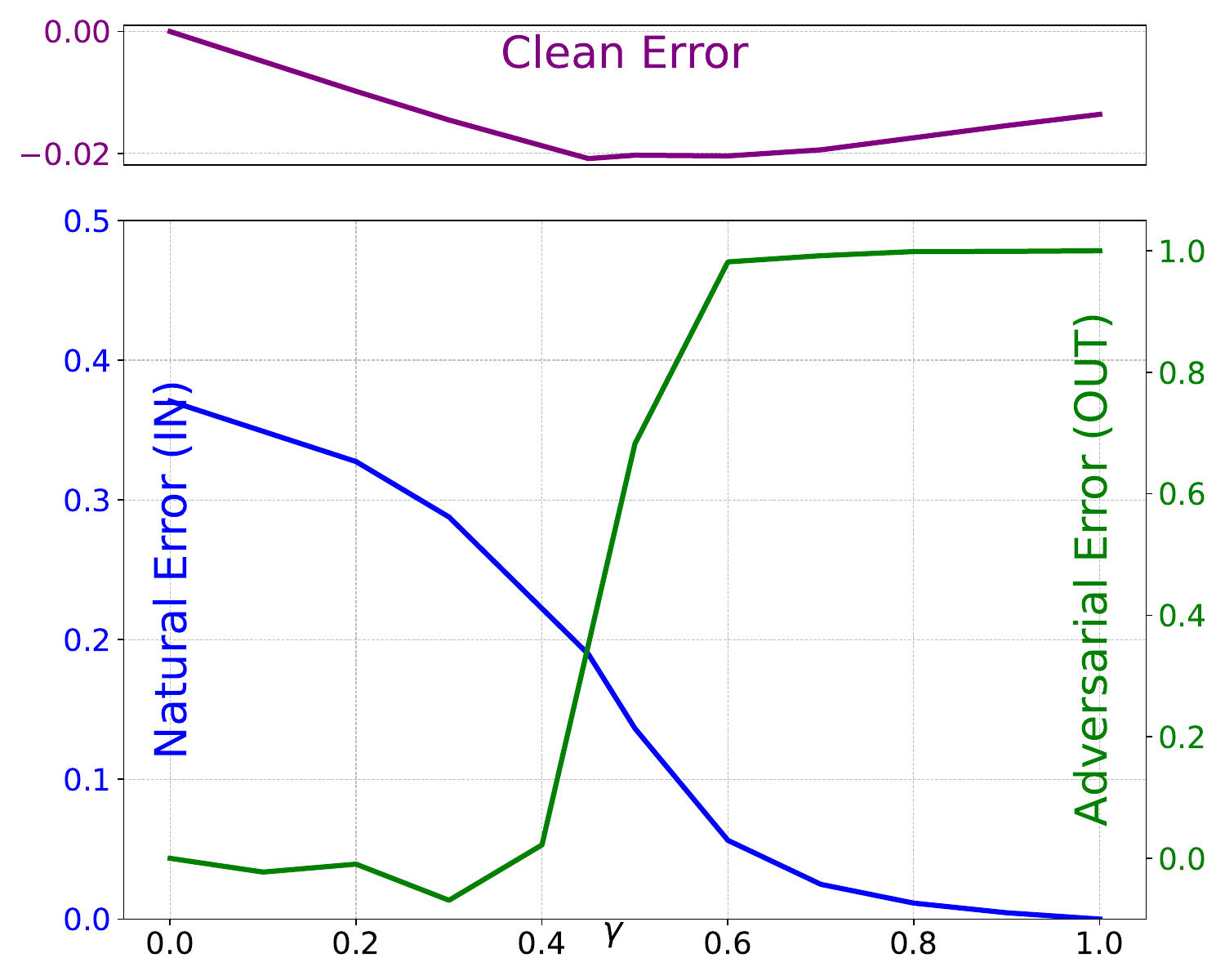}
            \caption{$f_1$: DeepLab\_MobileNet \\ $f_2$: SegFormer.\_b0}
        \end{subfigure}
        \begin{subfigure}{0.24\textwidth}
            \centering
            \includegraphics[width=\textwidth]{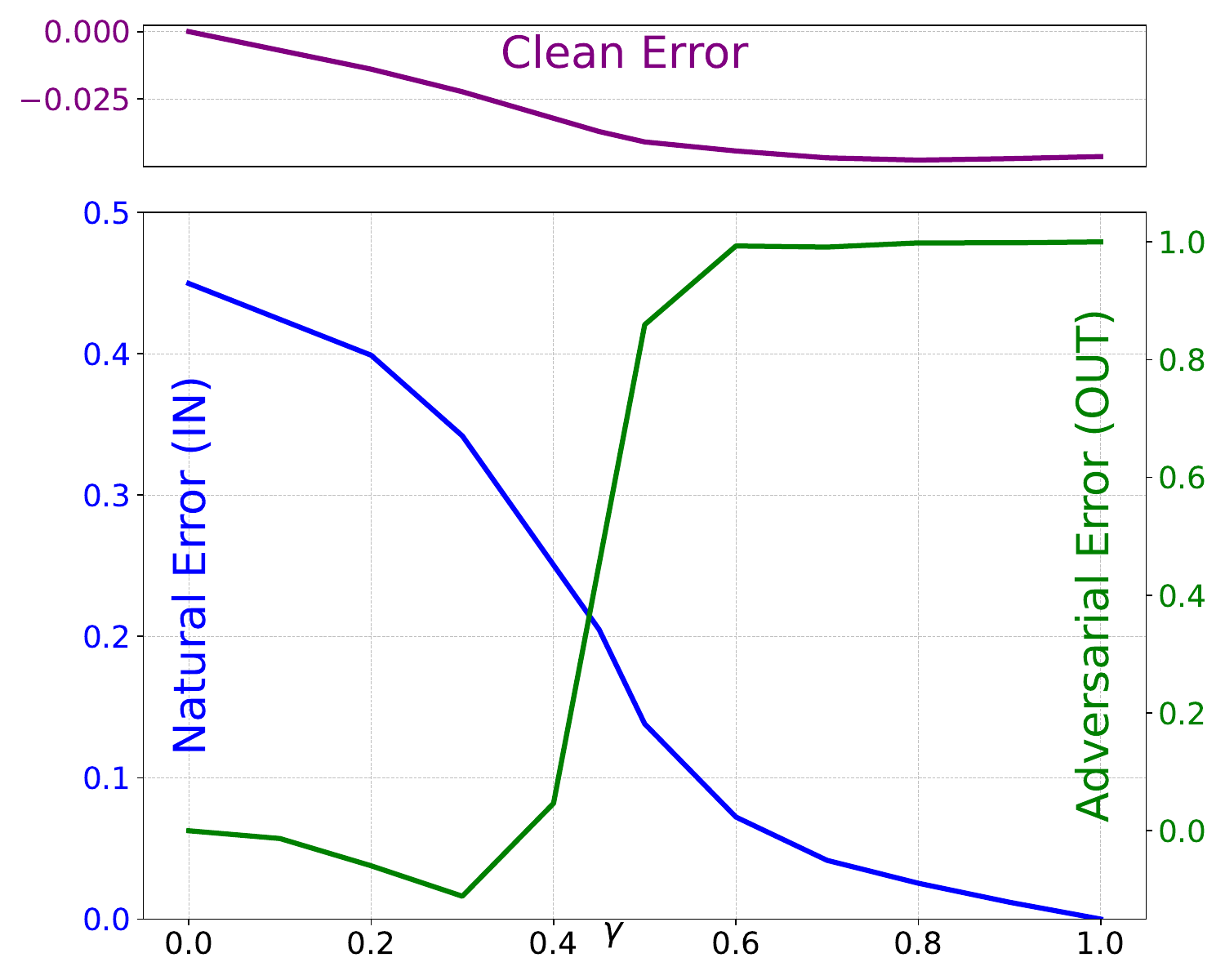}
            \caption{$f_1$: DeepLab\_MobileNet \\ $f_2$: SegFormer.\_b1}
        \end{subfigure}
         \begin{subfigure}{0.24\textwidth}
            \centering
            \includegraphics[width=\textwidth]{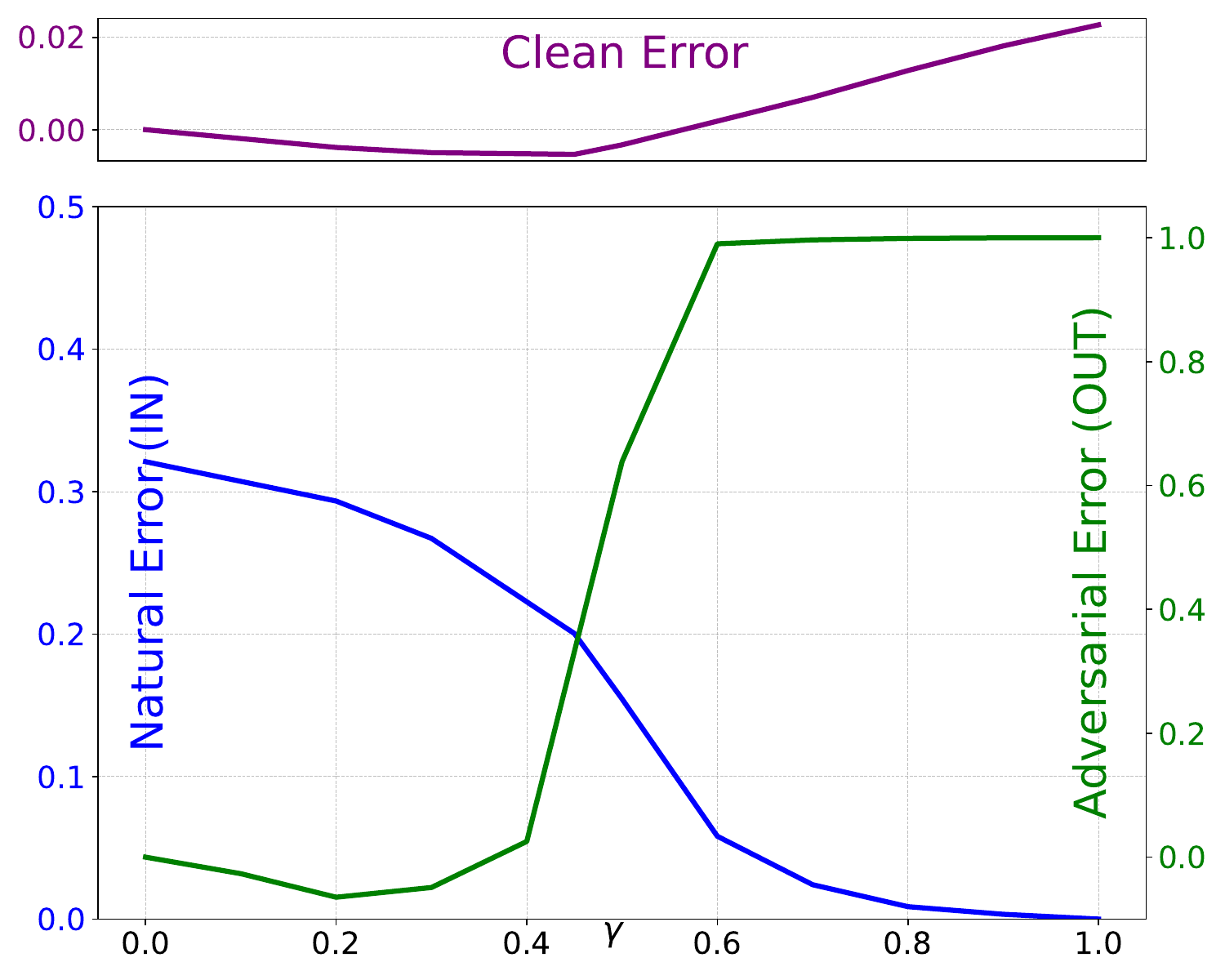}
            \caption{$f_1$: DeepLab\_ResNet \\ $f_2$: SegFormer.\_b0}
        \end{subfigure}
        \begin{subfigure}{0.24\textwidth}
            \centering
            \includegraphics[width=\textwidth]{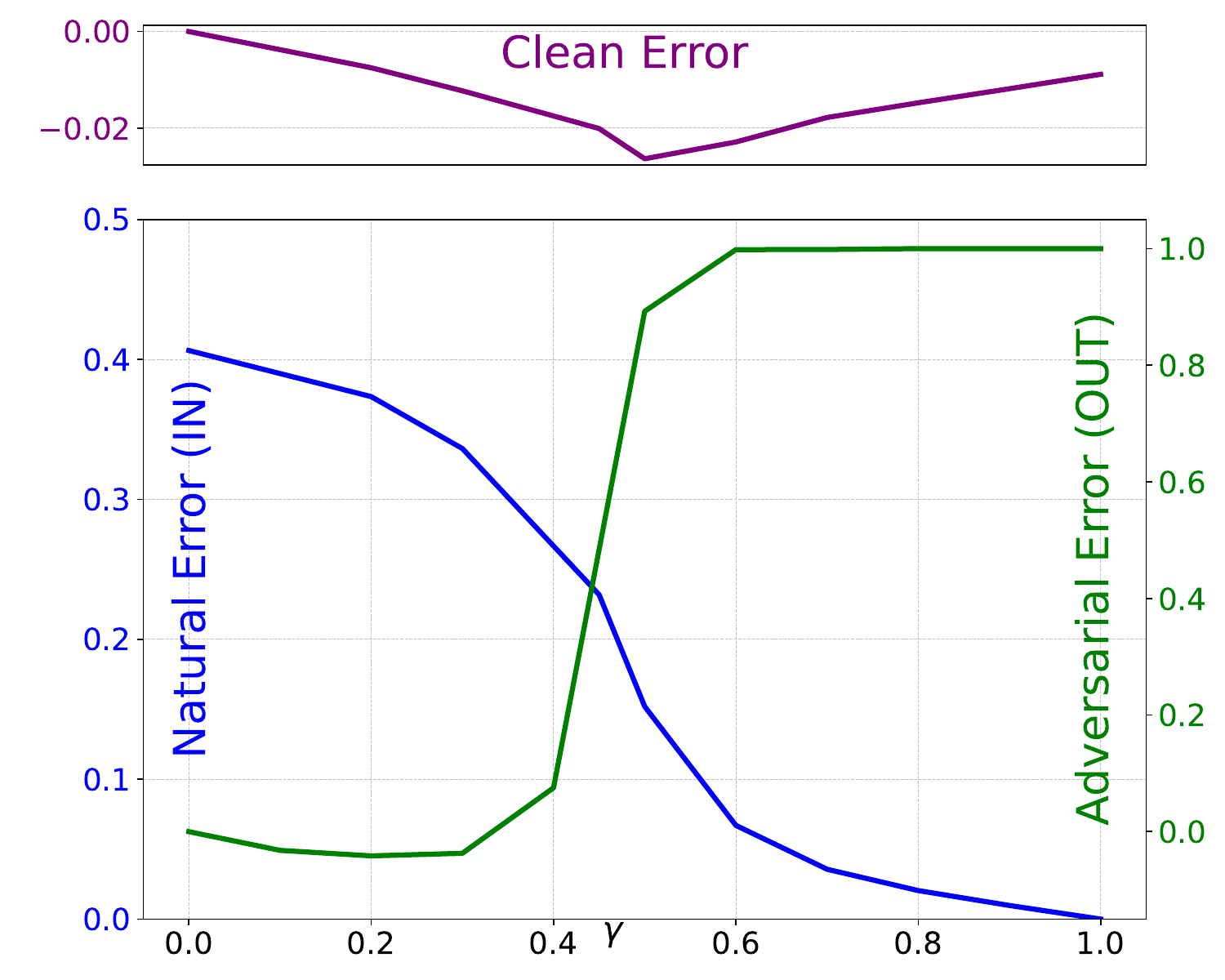}
            \caption{$f_1$: DeepLab\_ResNet \\ $f_2$: SegFormer.\_b1}
        \end{subfigure}
        \end{subfigure}
        \begin{subfigure}{\textwidth}
        \centering
        \begin{subfigure}{0.24\textwidth}
            \centering
            \includegraphics[width=\textwidth]{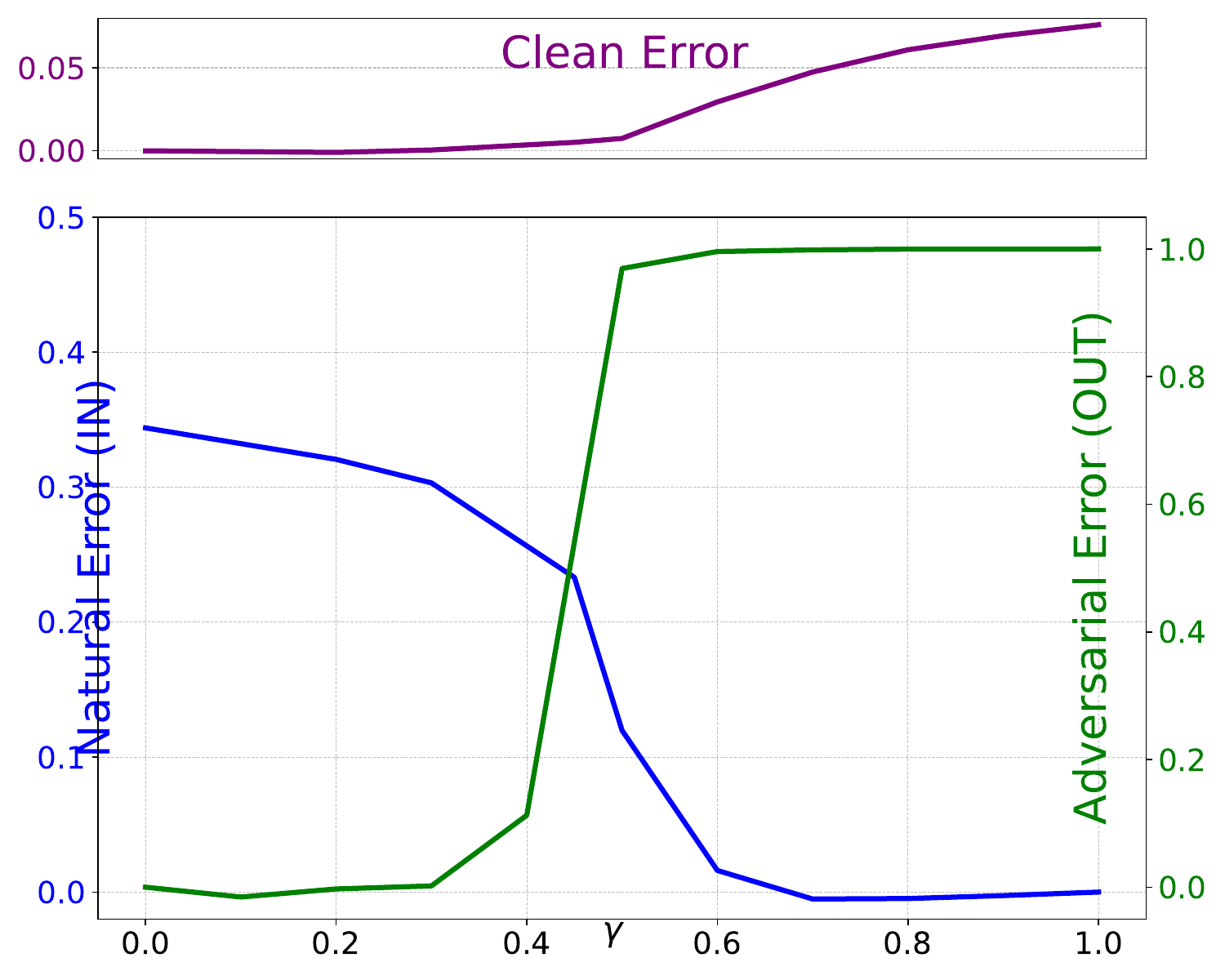}
            \caption{$f_1$: Bisenet-R101\\ $f_2$: SegFormer.\_b0}
        \end{subfigure}
        \begin{subfigure}{0.24\textwidth}
            \centering
            \includegraphics[width=\textwidth]{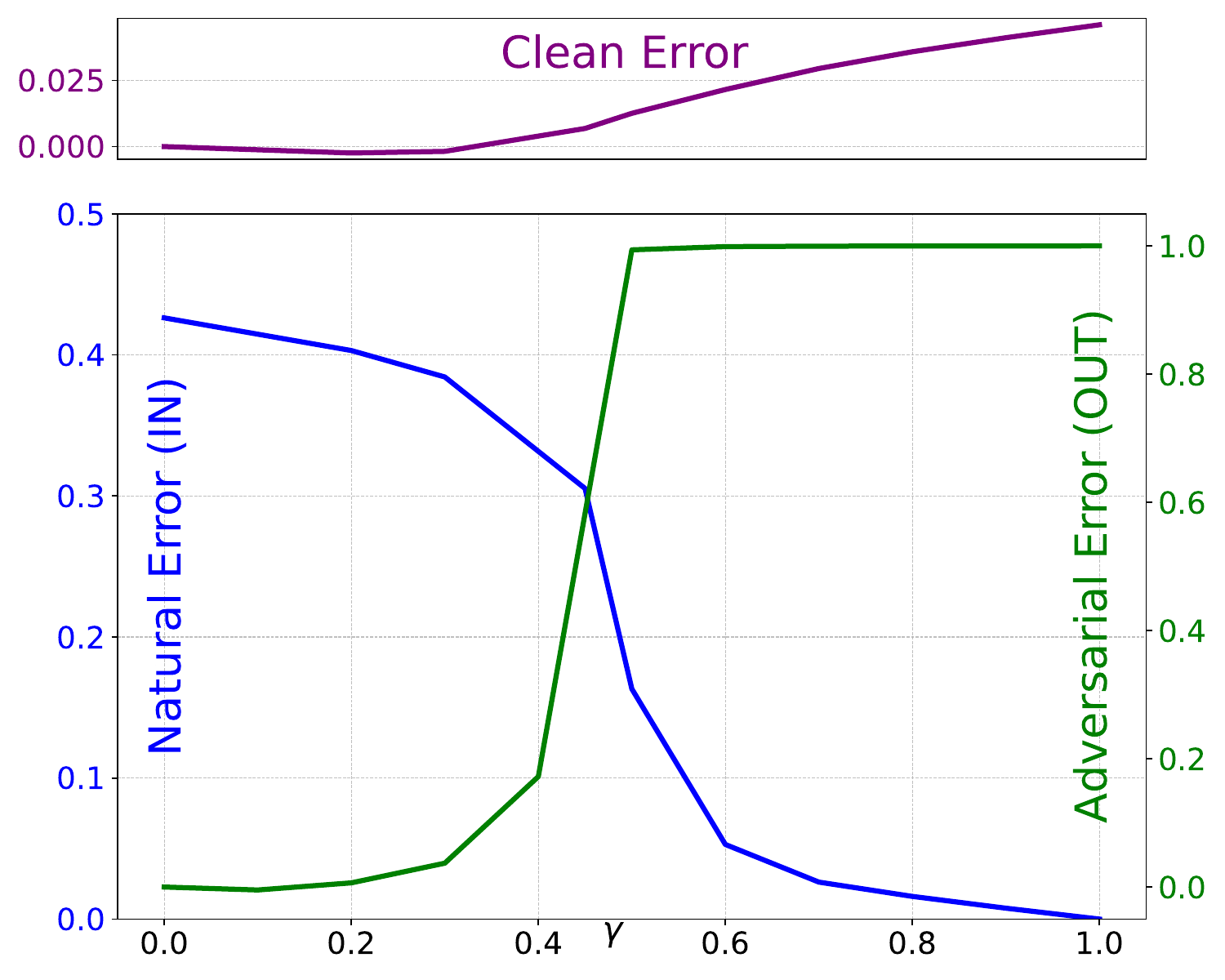}
            \caption{$f_1$: Bisenet-R101\\ $f_2$: SegFormer.\_b1}
        \end{subfigure}
         \begin{subfigure}{0.24\textwidth}
            \centering
            \includegraphics[width=\textwidth]{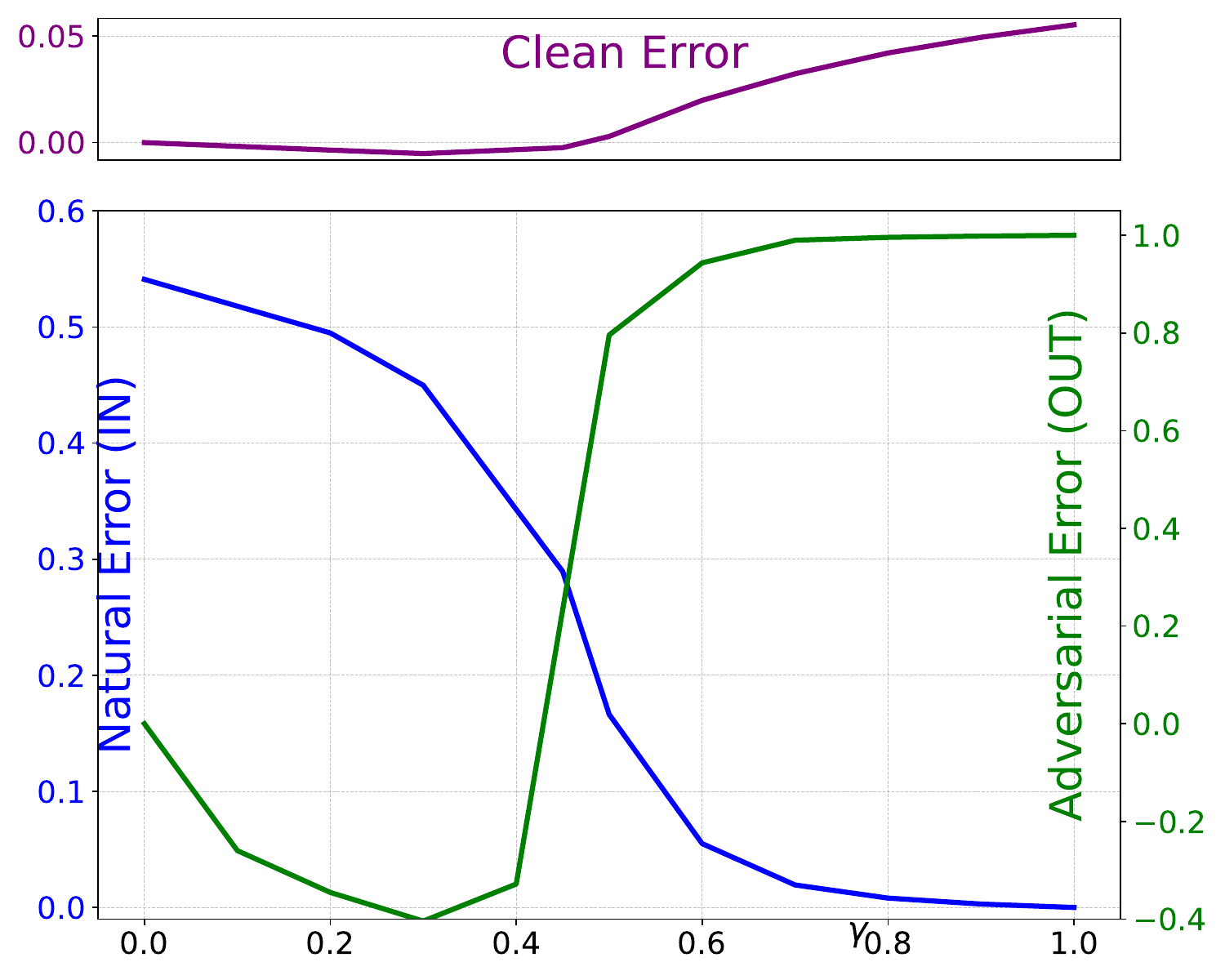}
            \caption{$f_1$: DDRNet23\\ $f_2$: SegFormer.\_b0}
        \end{subfigure}
        \begin{subfigure}{0.24\textwidth}
            \centering
            \includegraphics[width=\textwidth]{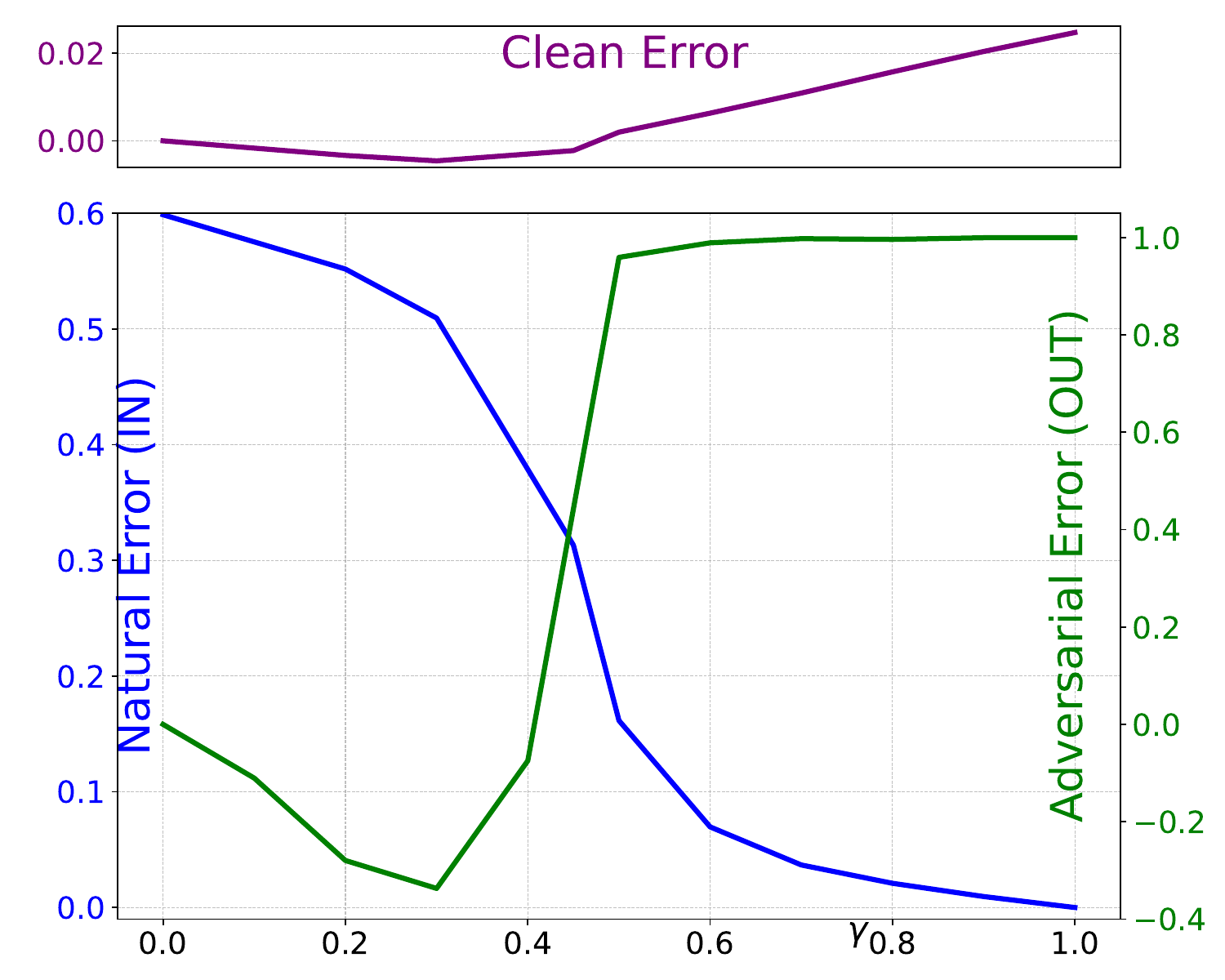}
            \caption{$f_1$:  DDRNet23 \\ $f_2$: SegFormer.\_b1}
        \end{subfigure}
        \end{subfigure}
        \caption{\captionEnsembleTest}
        \label{fig:enseble_analysis}
    \end{figure*}
\else
    \begin{figure*}[ht]
        \centering
        % First row of images
        \begin{subfigure}{\columnwidth}
        \begin{subfigure}{0.24\columnwidth}
            \centering
            \includegraphics[width=\textwidth]{ensembling_analysismobilenet_segb0-2.pdf}
            \caption{$f_1$: DeepLab\_MobileNet \\ $f_2$: SegFormer.\_b0}
        \end{subfigure}
        \begin{subfigure}{0.24\columnwidth}
            \centering
            \includegraphics[width=\textwidth]{ensembling_analysismobilenet_segb1-2.pdf}
            \caption{$f_1$: DeepLab\_MobileNet \\ $f_2$: SegFormer.\_b1}
        \end{subfigure}
         \begin{subfigure}{0.24\columnwidth}
            \centering
            \includegraphics[width=\textwidth]{ensembling_analysisresnet_segb0-2.pdf}
            \caption{$f_1$: DeepLab\_ResNet \\ $f_2$: SegFormer.\_b0}
        \end{subfigure}
        \begin{subfigure}{0.24\columnwidth}
            \centering
            \includegraphics[width=\textwidth]{ensembling_analysisresnet_segb1-2.pdf}
            \caption{$f_1$: DeepLab\_ResNet \\ $f_2$: SegFormer.\_b1}
        \end{subfigure}
        \end{subfigure}
        \begin{subfigure}{\columnwidth}
        \begin{subfigure}{0.24\columnwidth}
            \centering
            \includegraphics[width=\textwidth]{ensembling_analysisbisenet_segb0.pdf}
            \caption{$f_1$: Bisenet-R101\\ $f_2$: SegFormer.\_b0}
        \end{subfigure}
        \begin{subfigure}{0.24\columnwidth}
            \centering
            \includegraphics[width=\textwidth]{ensembling_analysisbisenet_segb1.pdf}
            \caption{$f_1$: Bisenet-R101\\ $f_2$: SegFormer.\_b1}
        \end{subfigure}
         \begin{subfigure}{0.24\columnwidth}
            \centering
            \includegraphics[width=\textwidth]{ensembling_analysisddrnet_segb0.pdf}
            \caption{$f_1$: DDRNet23\\ $f_2$: SegFormer.\_b0}
        \end{subfigure}
        \begin{subfigure}{0.24\columnwidth}
            \centering
            \includegraphics[width=\textwidth]{ensembling_analysisddrnet_segb1.pdf}
            \caption{$f_1$:  DDRNet23 \\ $f_2$: SegFormer.\_b1}
        \end{subfigure}
        \end{subfigure}
        \caption{\captionEnsembleTest}
        \label{fig:enseble_analysis}
    \end{figure*}
\fi
\makeatother

%% file: conclusion.tex
\section{Discussion and Conclusion}
\label{s:conclusion}
This work presented an extensive analysis of the robustness of segmentation models under localized corruptions, considering both natural and adversarial \GM{corruptions}. We first introduced metrics that extend the classic pixel-wise accuracy, commonly used in segmentation, to provide a more comprehensive analysis of spatial robustness, both within and beyond the corrupted regions. We then proposed a framework that enables a systematic evaluation of localized natural and adversarial corruptions by allowing the selection of parameters, including corruption ratios and model patch sizes.

Our analysis of adversarial perturbations also improves upon existing strategies in the literature by demonstrating that, in dense prediction tasks such as semantic segmentation, targeting multiple pixels introduces a complex multi-objective problem. This complexity makes it difficult to assess \GM{robustness in} the worst-case adversarial scenario using a single localized perturbation. \GM{Therefore, we proposed the \textit{region-aware multi-attack adversarial analysis}, a method to assess the spatial robustness of models by applying multiple $\infty$-norm attacks while continuously adjusting the fooling area.} % thus providing a more realistic assessment of model robustness across different parts of the output.}
%To address this, we proposed a \textit{Region-aware multi-attack adversarial analysis}, a method that applies multiple attacks while continuously adjusting the fooling area, thus providing a more realistic assessment of model robustness across different parts of the output.

\GM{All findings were evaluated across a diverse set of segmentation models in a driving scenario, specifically using the Cityscapes dataset \cite{cordts2016cityscapes}.} Our results revealed distinct differences in model behavior under natural and adversarial corruptions. Furthermore, we highlighted the importance of finding a trade-off between these two robustness aspects, which are not necessarily correlated. As a preliminary step toward addressing this challenge, we explored ensemble strategies that combine convolution-based models with transformer-based architectures.

\GM{
In a future work, we aim at further investigating this trade-off by exploring more complex ensembling approaches, extending the analysis to other tasks and types or forms of corruptions, and providing theoretical insights into this intriguing mixed property. Another important direction is to integrate the proposed localized robustness analysis into model training strategies through the use of localized augmentation techniques (e.g., Cutout \cite{devries2017improved_cutout} and CutMix \cite{yun2019cutmix}), which are rarely applied in complex scene understanding tasks such as semantic segmentation. This could provide valuable insights into the development of novel defense mechanisms and robust training approaches. In this work, we focused on standalone models trained with conventional image-level augmentations, leaving this integration as a promising direction for future research.
}

To conclude, our study represents an important advancement in understanding the spatial robustness of segmentation models, while also identifying key challenges and open research directions for future investigations.

{\small
\section*{Acknowledgements}
This work was partially supported by project SERICS (PE00000014) under the MUR (Ministero dell'Università e della Ricerca) National Recovery and Resilience Plan funded by the European Union - NextGenerationEU.}

%% file: egbib.bib
@article{pintor_imagenetpatch,
title = {ImageNet-Patch: A dataset for benchmarking machine learning robustness against adversarial patches},
journal = {Pattern Recognition},
volume = {134},
pages = {109064},
year = {2023},
issn = {0031-3203},
url = {https://www.sciencedirect.com/science/article/pii/S0031320322005441},
author = {Maura Pintor and Daniele Angioni and Angelo Sotgiu and Luca Demetrio and Ambra Demontis and Battista Biggio and Fabio Roli},
}

@article{liu2025_patch,
title = {RADAP: A Robust and Adaptive Defense Against Diverse Adversarial Patches on face recognition},
journal = {Pattern Recognition},
volume = {157},
pages = {110915},
year = {2025},
issn = {0031-3203},
doi = {https://doi.org/10.1016/j.patcog.2024.110915},
url = {https://www.sciencedirect.com/science/article/pii/S0031320324006666},
author = {Xiaoliang Liu and Furao Shen and Jian Zhao and Changhai Nie},
}

@article{chen2017rethinking_deeplab,
  title={Rethinking atrous convolution for semantic image segmentation},
  author={Chen, Liang-Chieh and Papandreou, George and Schroff, Florian and Adam, Hartwig},
  journal={arXiv preprint arXiv:1706.05587},
  year={2017}
}

@article{xie2021segformer,
  title={SegFormer: Simple and efficient design for semantic segmentation with transformers},
  author={Xie, Enze and Wang, Wenhai and Yu, Zhiding and Anandkumar, Anima and Alvarez, Jose M and Luo, Ping},
  journal={Advances in neural information processing systems},
  volume={34},
  pages={12077--12090},
  year={2021}
}

@inproceedings{xu2023pidnet,
  title={PIDNet: A real-time semantic segmentation network inspired by PID controllers},
  author={Xu, Jiacong and Xiong, Zixiang and Bhattacharyya, Shankar P},
  booktitle={Proceedings of the IEEE/CVF conference on computer vision and pattern recognition},
  pages={19529--19539},
  year={2023}
}

@inproceedings{yun2019cutmix,
  title={Cutmix: Regularization strategy to train strong classifiers with localizable features},
  author={Yun, Sangdoo and Han, Dongyoon and Oh, Seong Joon and Chun, Sanghyuk and Choe, Junsuk and Yoo, Youngjoon},
  booktitle={Proceedings of the IEEE/CVF international conference on computer vision},
  pages={6023--6032},
  year={2019}
}

@article{devries2017improved_cutout,
  title={Improved regularization of convolutional neural networks with cutout},
  author={DeVries, Terrance and Taylor, Graham W},
  journal={arXiv preprint arXiv:1708.04552},
  year={2017}
}

@article{luo2016understanding,
  title={Understanding the effective receptive field in deep convolutional neural networks},
  author={Luo, Wenjie and Li, Yujia and Urtasun, Raquel and Zemel, Richard},
  journal={Advances in neural information processing systems},
  volume={29},
  year={2016}
}

@inproceedings{metzen2021meta,
	title        = {Meta adversarial training against universal patches},
	author       = {Metzen, Jan Hendrik and Finnie, Nicole and Hutmacher, Robin},
	year         = 2021,
	booktitle    = {ICML 2021 Workshop on Adversarial Machine Learning}
}

@article{rossolini_tnnls_2023,
	title        = {On the Real-World Adversarial Robustness of Real-Time Semantic Segmentation Models for Autonomous Driving},
	author       = {Rossolini, Giulio and Nesti, Federico and D’Amico, Gianluca and Nair, Saasha and Biondi, Alessandro and Buttazzo, Giorgio},
	year         = 2023,
	journal      = {IEEE Transactions on Neural Networks and Learning Systems},
	volume       = {},
	number       = {},
	pages        = {1--15},
	doi          = {10.1109/TNNLS.2023.3314512}
}

@inproceedings{biggio2018wild,
  title={Wild patterns: Ten years after the rise of adversarial machine learning},
  author={Biggio, Battista and Roli, Fabio},
  booktitle={Proceedings of the 2018 ACM SIGSAC Conference on Computer and Communications Security},
  pages={2154--2156},
  year={2018}
}

@article{brown_adversarial_2018,
	title        = {Adversarial {Patch}},
	author       = {Brown, Tom B. and Mané, Dandelion and Roy, Aurko and Abadi, Martín and Gilmer, Justin},
	year         = 2018,
	month        = may,
	journal      = {arXiv:1712.09665 [cs]},
	urldate      = {2021-05-13}
}

@inproceedings{arnab_robustness_nodate,
	title        = {On the robustness of semantic segmentation models to adversarial attacks},
	author       = {Arnab, Anurag and Miksik, Ondrej and Torr, Philip HS},
	year         = 2018,
	booktitle    = {Proceedings of the IEEE Conference on Computer Vision and Pattern Recognition},
	pages        = {888--897}
}

@inproceedings{nakka_indirect_2020,
	title        = {Indirect Local Attacks for Context-Aware Semantic Segmentation Networks},
	author       = {Krishna Kanth Nakka and Mathieu Salzmann},
	year         = 2020,
	booktitle    = {16th European Conference Computer Vision {ECCV}},
	publisher    = {Springer},
	volume       = 12350,
	pages        = {}
}

@inproceedings{Szegedy14,
	title        = {Intriguing properties of neural networks},
	author       = {Christian Szegedy and Wojciech Zaremba and Ilya Sutskever and Joan Bruna and Dumitru Erhan and Ian J. Goodfellow and Rob Fergus},
	year         = 2014,
	booktitle    = {2nd International Conference on Learning Representations, {ICLR}}
}

@inproceedings{Carlini017,
	title        = {Towards Evaluating the Robustness of Neural Networks},
	author       = {Nicholas Carlini and David A. Wagner},
	year         = 2017,
	booktitle    = {2017 {IEEE} Symposium on Security and Privacy, {SP} 2017, San Jose, CA, USA, May 22-26, 2017},
	publisher    = {{IEEE} Computer Society},
	pages        = {39--57},
	doi          = {10.1109/SP.2017.49},
	url          = {https://doi.org/10.1109/SP.2017.49},
	bibsource    = {dblp computer science bibliography, https://dblp.org}
}

@article{survey_trust,
	title        = {A survey of safety and trustworthiness of deep neural networks: Verification, testing, adversarial attack and defence, and interpretability},
	author       = {Huang, Xiaowei and Kroening, Daniel and Ruan, Wenjie and Sharp, James and Sun, Youcheng and Thamo, Emese and Wu, Min and Yi, Xinping},
	year         = 2020,
	journal      = {Computer Science Review},
	publisher    = {Elsevier},
	volume       = 37,
	pages        = 100270
}

@inproceedings{DBLP:conf/cvpr/CordtsORREBFRS16,
	title        = {The Cityscapes Dataset for Semantic Urban Scene Understanding},
	author       = {Marius Cordts and Mohamed Omran and Sebastian Ramos and Timo Rehfeld and Markus Enzweiler and Rodrigo Benenson and Uwe Franke and Stefan Roth and Bernt Schiele},
	year         = 2016,
	booktitle    = {Conference on Computer Vision and Pattern Recognition {CVPR}},
	publisher    = {{IEEE} Computer Society},
	pages        = {3213--3223},
	bibsource    = {dblp computer science bibliography, https://dblp.org}
}

@article{ddrnet_paper,
	title        = {Deep dual-resolution networks for real-time and accurate semantic segmentation of road scenes},
	author       = {Hong, Yuanduo and Pan, Huihui and Sun, Weichao and Jia, Yisong},
	year         = 2021,
	journal      = {arXiv:2101.06085}
}

@inproceedings{icnet_paper,
	title        = {Icnet for real-time semantic segmentation on high-resolution images},
	author       = {Zhao, Hengshuang and Qi, Xiaojuan and Shen, Xiaoyong and Shi, Jianping and Jia, Jiaya},
	year         = 2018,
	booktitle    = {European Conference on Computer Vision (ECCV)},
	publisher    = {Springer},
	pages        = {405--420}
}

@inproceedings{bisenet_paper,
	title        = {Bisenet: Bilateral segmentation network for real-time semantic segmentation},
	author       = {Yu, Changqian and Wang, Jingbo and Peng, Chao and Gao, Changxin and Yu, Gang and Sang, Nong},
	year         = 2018,
	booktitle    = {European Conference on Computer Vision},
	publisher    = {Springer},
	pages        = {}
}

@inproceedings{pspnet_paper,
	title        = {Pyramid Scene Parsing Network},
	author       = {Zhao, Hengshuang and Shi, Jianping and Qi, Xiaojuan and Wang, Xiaogang and Jia, Jiaya},
	year         = 2017,
	booktitle    = {2017 IEEE Conference on Computer Vision and Pattern Recognition (CVPR)},
	volume       = {},
	number       = {},
	pages        = {6230--6239},
	doi          = {10.1109/CVPR.2017.660}
}

@inproceedings{pgd_attack,
	title        = {Towards Deep Learning Models Resistant to Adversarial Attacks},
	author       = {Aleksander Madry and Aleksandar Makelov and Ludwig Schmidt and Dimitris Tsipras and Adrian Vladu},
	year         = 2018,
	booktitle    = {6th International Conference on Learning Representations, {ICLR} 2018, Vancouver, BC, Canada, April 30 - May 3, 2018, Conference Track Proceedings},
	publisher    = {OpenReview.net},
	url          = {https://openreview.net/forum?id=rJzIBfZAb},
	bibsource    = {dblp computer science bibliography, https://dblp.org}
}

@inproceedings{cordts2016cityscapes,
	title        = {The cityscapes dataset for semantic urban scene understanding},
	author       = {Cordts, Marius and Omran, Mohamed and Ramos, Sebastian and Rehfeld, Timo and Enzweiler, Markus and Benenson, Rodrigo and Franke, Uwe and Roth, Stefan and Schiele, Bernt},
	year         = 2016,
	booktitle    = {Proceedings of the IEEE conference on computer vision and pattern recognition},
	pages        = {}
}

@inproceedings{saha_role_2020,
	title        = {Role of Spatial Context in Adversarial Robustness for Object Detection},
	author       = {Saha, Aniruddha and Subramanya, Akshayvarun and Patil, Koninika and Pirsiavash, Hamed},
	year         = 2020,
	booktitle    = {2020 {IEEE}/{CVF} Conference on Computer Vision and Pattern Recognition Workshops ({CVPRW})},
	location     = {Seattle, {WA}, {USA}},
	publisher    = {{IEEE}},
	pages        = {3403--3412}
}

@inproceedings{vaswani_attention_2017,
	title        = {Attention is All you Need},
	author       = {Vaswani, Ashish and Shazeer, Noam and Parmar, Niki and Uszkoreit, Jakob and Jones, Llion and Gomez, Aidan N and Kaiser, {\L}ukasz and Polosukhin, Illia},
	booktitle    = {Advances in Neural Information Processing Systems},
	publisher    = {Curran Associates, Inc.},
	volume       = 30,
	url          = {https://proceedings.neurips.cc/paper/2017/hash/3f5ee243547dee91fbd053c1c4a845aa-Abstract.html},
	urldate      = {2021-09-21},
	year         = 2017
}

@article{patch_object,
	title        = {On Physical Adversarial Patches for Object Detection},
	author       = {Mark Lee and J. Zico Kolter},
	year         = 2019,
	journal      = {CoRR},
	volume       = {abs/1906.11897},
	url          = {},
	eprinttype   = {arXiv},
	eprint       = {},
	bibsource    = {}
}

@article{brau2022minimal,
	title        = {On the Minimal Adversarial Perturbation for Deep Neural Networks With Provable Estimation Error},
	author       = {Brau, Fabio and Rossolini, Giulio and Biondi, Alessandro and Buttazzo, Giorgio},
	year         = 2022,
	journal      = {IEEE Transactions on Pattern Analysis and Machine Intelligence},
	volume       = {},
	number       = {},
	pages        = {1--15},
	doi          = {10.1109/TPAMI.2022.3195616}
}

@INPROCEEDINGS{rossolini_iccps24,
  author={Rossolini, Giulio and Biondi, Alessandro and Buttazzo, Giorgio},
  booktitle={2024 ACM/IEEE 15th International Conference on Cyber-Physical Systems (ICCPS)}, 
  title={Attention-Based Real-Time Defenses for Physical Adversarial Attacks in Vision Applications}, 
  year={2024},
  volume={},
  number={},
  pages={23-32},
  doi={10.1109/ICCPS61052.2024.00009}}

@article{yuan2024towards,
  title={Towards Robust Semantic Segmentation against Patch-Based Attack via Attention Refinement},
  author={Yuan, Zheng and Zhang, Jie and Wang, Yude and Shan, Shiguang and Chen, Xilin},
  journal={International Journal of Computer Vision},
  pages={1--23},
  year={2024},
  publisher={Springer}
}

@article{ensemble_GANAIE2022105151,
title = {Ensemble deep learning: A review},
journal = {Engineering Applications of Artificial Intelligence},
volume = {115},
pages = {105151},
year = {2022},
issn = {0952-1976},
doi = {https://doi.org/10.1016/j.engappai.2022.105151},
author = {M.A. Ganaie and Minghui Hu and A.K. Malik and M. Tanveer and P.N. Suganthan},
}

@inproceedings{hendrycks2018benchmarking,
  title={Benchmarking Neural Network Robustness to Common Corruptions and Perturbations},
  author={Hendrycks, Dan and Dietterich, Thomas},
  booktitle={International Conference on Learning Representations},
  year={2018}
}

@article{croce2022interplay,
  title={On the interplay of adversarial robustness and architecture components: patches, convolution and attention},
  author={Croce, Francesco and Hein, Matthias},
  journal={arXiv preprint arXiv:2209.06953},
  year={2022}
}

@inproceedings{kamann2020benchmarking,
  title={Benchmarking the robustness of semantic segmentation models},
  author={Kamann, Christoph and Rother, Carsten},
  booktitle={Proceedings of the IEEE/CVF conference on computer vision and pattern recognition},
  pages={8828--8838},
  year={2020}
}

@inproceedings{schiappa2024robustness,
  title={Robustness Analysis on Foundational Segmentation Models},
  author={Schiappa, Madeline Chantry and Azad, Shehreen and Vs, Sachidanand and Ge, Yunhao and Miksik, Ondrej and Rawat, Yogesh S and Vineet, Vibhav},
  booktitle={Proceedings of the IEEE/CVF Conference on Computer Vision and Pattern Recognition},
  pages={1786--1796},
  year={2024}
}

@inproceedings{gu2022vision,
  title={Are vision transformers robust to patch perturbations?},
  author={Gu, Jindong and Tresp, Volker and Qin, Yao},
  booktitle={European Conference on Computer Vision},
  pages={404--421},
  year={2022},
  organization={Springer}
}

@inproceedings{bhojanapalli2021understanding,
  title={Understanding robustness of transformers for image classification},
  author={Bhojanapalli, Srinadh and Chakrabarti, Ayan and Glasner, Daniel and Li, Daliang and Unterthiner, Thomas and Veit, Andreas},
  booktitle={Proceedings of the IEEE/CVF international conference on computer vision},
  pages={10231--10241},
  year={2021}
}

@article{naseer2021intriguing,
  title={Intriguing properties of vision transformers},
  author={Naseer, Muhammad Muzammal and Ranasinghe, Kanchana and Khan, Salman H and Hayat, Munawar and Shahbaz Khan, Fahad and Yang, Ming-Hsuan},
  journal={Advances in Neural Information Processing Systems},
  volume={34},
  pages={23296--23308},
  year={2021}
}

@inproceedings{mahmood2021robustness,
  title={On the robustness of vision transformers to adversarial examples},
  author={Mahmood, Kaleel and Mahmood, Rigel and Van Dijk, Marten},
  booktitle={Proceedings of the IEEE/CVF international conference on computer vision},
  pages={7838--7847},
  year={2021}
}

@article{survey_liu2024survey,
  title={A survey on autonomous driving datasets: Statistics, annotation quality, and a future outlook},
  author={Liu, Mingyu and Yurtsever, Ekim and Fossaert, Jonathan and Zhou, Xingcheng and Zimmer, Walter and Cui, Yuning and Zagar, Bare Luka and Knoll, Alois C},
  journal={IEEE Transactions on Intelligent Vehicles},
  year={2024},
  publisher={IEEE}
}

@article{PR_QIAN2022108796,
title = {3D Object Detection for Autonomous Driving: A Survey},
journal = {Pattern Recognition},
volume = {130},
pages = {108796},
year = {2022},
issn = {0031-3203},
doi = {https://doi.org/10.1016/j.patcog.2022.108796},
author = {Rui Qian and Xin Lai and Xirong Li},
}
